\pdfoutput=1

\documentclass[11pt]{article}

\usepackage[]{ACL2023}

\usepackage{times}
\usepackage{latexsym}

\usepackage[T1]{fontenc}

\usepackage[utf8]{inputenc}

\usepackage{microtype}

\usepackage{inconsolata}

\usepackage{multirow}
\usepackage{color, colortbl}

\usepackage{microtype}
\usepackage{booktabs} 
\usepackage{mathtools}
\usepackage{tikz}
\usetikzlibrary{bayesnet}
\usetikzlibrary{arrows}
\usepackage{tabularx}
\usepackage{tabulary}
\usepackage{colortbl}
\usepackage{xspace}
\usepackage{float}

\usepackage[noabbrev,capitalise]{cleveref}
\crefname{table}{Table}{}
\crefname{figure}{Figure}{}
\crefname{algorithm}{Algorithm}{}
\crefname{equation}{Eq.}{}
\crefname{appendix}{App.}{}
\crefname{prop}{Proposition}{}
\crefname{thm}{Theorem}{}

\newcommand\doublecheck{{\checked\kern-0.5em\checked}}
\usepackage{fontawesome5}
\newcommand\checked{\footnotesize \faIcon{check}}

\usepackage{mathtools}
\newcounter{tableeqn}[table]

\newcounter{tablesubeqn}[tableeqn]

\usepackage{relsize}
\usepackage{amsmath, amsthm, amssymb}
\usepackage{mathrsfs}

\usepackage{caption}
\usepackage{graphicx}

\usepackage{enumitem}

\usepackage{color, colortbl}
\PassOptionsToPackage{table,dvipsnames}{xcolor}
\usepackage{multirow,makecell}
\usepackage{rotating}
\usepackage{booktabs}
\usepackage[table]{xcolor}
\usepackage{tabularx}

\usepackage{todonotes}
\makeatletter
\newcommand*\iftodonotes{\if@todonotes@disabled\expandafter\@secondoftwo\else\expandafter\@firstoftwo\fi}
\makeatother

\newcommand{\Table}[1]{Table~\ref{#1}}

\newcommand{\ignore}[1]{}

\newcommand\ie{\emph{i.e.}}
\newcommand\eg{\emph{e.g.}}

\newcommand{\vl}{V\&L\xspace}

\newcommand{\gray}[1]{\textcolor{gray}{#1}}

\newcommand{\albef}{{\sc ALBEF}\xspace}
\newcommand{\albefb}{{\sc ALBEF$_\text{4M}$}\xspace}
\newcommand{\albefl}{{\sc ALBEF$_\text{14M}$}\xspace}
\newcommand{\blip}{{\sc BLIP}\xspace}
\newcommand{\blipl}{{\sc BLIP$_\text{14M}$}\xspace}
\newcommand{\blipxl}{{\sc BLIP$_\text{129M}$}\xspace}
\newcommand{\blipxlfilt}{{\sc BLIP$_\text{129M}$-CapFilt/L}\xspace}
\newcommand{\blipvitxl}{{\sc BLIP-ViT/L$_\text{129M}$}\xspace}
\newcommand{\pevl}{{\sc PEVL}\xspace}
\newcommand{\pevll}{{\sc PEVL$_\text{14M}$}\xspace}
\newcommand{\pevlpre}{{\sc PEVL$_\text{14M}$}\xspace}
\newcommand{\pevlgrd}{{\sc PEVL$_\text{grd}$}\xspace}

\newcommand{\pevlvrd}{{\sc PEVL$_\text{vrd}$}\xspace}
\newcommand{\xvlm}{{\sc X-VLM}\xspace}
\newcommand{\xvlmb}{{\sc X-VLM$_\text{4M}$}\xspace}
\newcommand{\xvlmm}{{\sc X-VLM$_\text{14M}$}\xspace}
\newcommand{\xvlml}{{\sc X-VLM$_\text{16M}$}\xspace}
\newcommand{\capfilt}{{\sc CapFilt}\xspace}
\newcommand{\capfiltb}{{\sc CapFilt/B}\xspace}
\newcommand{\capfiltl}{{\sc CapFilt/L}\xspace}

\newcommand{\clip}{{\sc CLIP}\xspace}
\newcommand{\clipcap}{{\sc ClipCap}\xspace}
\newcommand{\flamingo}{{\sc Flamingo}\xspace}
\newcommand{\bliptwo}{{\sc BLIP-2}\xspace}

\newcommand{\vgrd}{{\sc VG$_\text{RD}$}\xspace}
\newcommand{\vgod}{{\sc VG$_\text{OD}$}\xspace}
\newcommand{\cocod}{{\sc COCO$_\text{OD}$}\xspace}

\newcommand{\lvma}{{$\mathcal{L}_\text{VMA}$}\xspace}
\newcommand{\lbbox}{{$\mathcal{L}_\text{bbox}$}\xspace}

\title{Measuring Progress in Fine-grained Vision-and-Language Understanding}

\usepackage{tipa}
\newcommand{\deepmind}{\text{\normalfont \textipa{D}}}
\newcommand{\ku}{\text{\normalfont \textipa{C}}}

\author{Emanuele Bugliarello\Thanks{Work completed during an internship at DeepMind. $^\ddagger$denotes equal senior contribution. Correspondence to: Emanuele Bugliarello <\href{mailto:emanuele@di.ku.dk}{emanuele@di.ku.dk}>.}$^{~,\deepmind,\ku}$ \ \ 
        Laurent Sartran$^{\deepmind}$ \ \  
        Aishwarya Agrawal$^{\deepmind}$ \ \
        \\{\bf Lisa Anne Hendricks$^{\ddagger,\deepmind}$ \ \ Aida Nematzadeh$^{\ddagger,\deepmind}$} \\ \\
        $^{\deepmind}$DeepMind \ \ $^{\ku}$University of Copenhagen\ \ 
}

\begin{document}
\maketitle
\vspace*{-10mm} 
\begin{abstract}
    While pretraining on large-scale image–text data from the Web has facilitated rapid progress on many vision-and-language (\vl) tasks, recent work has demonstrated that pretrained models lack ``fine-grained'' understanding, such as the ability to recognise relationships, verbs, and numbers in images. This has resulted in an increased interest in the community to either develop new benchmarks or models for such capabilities. To better understand and quantify progress in this direction, we investigate four competitive \vl models on four fine-grained benchmarks. Through our analysis, we find that X-VLM~\cite{x-vlm} consistently outperforms other baselines, and that modelling innovations can impact performance more than scaling Web data, which even degrades performance sometimes. Through a deeper investigation of X-VLM, we highlight the importance of both novel losses and rich data sources for learning fine-grained skills. Finally, we inspect training dynamics, and discover that for some tasks, performance peaks early in training or significantly fluctuates, never converging.
\end{abstract}

\section{Introduction}

\emph{Fine-grained} multimodal skills (\eg, understanding relationships and  recognising verbs) require identifying and relating various entities across both image and text modalities.
Vision-and-language models (VLMs) need such skills to 
robustly perform well on real-world vision-and-language (\vl) applications; \eg, a \emph{coarse-grained} model tested on image retrieval to ``find an image where something is \emph{on} a sofa'' might incorrectly return an image of a cat sitting \emph{below} the sofa. As another example, in captioning, a model might incorrectly describe an image where ``someone is \emph{selling} a sweater'' as ``someone is \emph{buying} a sweater,'' if it does not have a precise understanding of the two verbs.

However, common V\&L benchmarks \citep[\eg,][]{coco,vqav2,nlvr2} do not explicitly shed light on such fine-grained understanding. 
Indeed, in the last few years, there has been an increase in the number of benchmarks which demonstrate that current, coarse-grained models struggle with fine-grained understanding~\cite{svo_probes, valse, probing_vl, winoground}.
Meanwhile, more models have been designed specifically to learn a better mapping between visual and textual modalities~\citep[\eg,][]{filip,pevl,x-vlm,pyramidclip}.
While such models perform well on coarse-grained retrieval and other downstream tasks, they have not been directly evaluated on fine-grained understanding.
Consequently, it is unclear if the performance gains are due to tighter, more fine-grained representations introduced by model innovations at the pretraining stage.
To fill this gap, we analyse several recent models with innovations designed for a better image--text alignment and their corresponding baselines on a suite of fine-grained benchmarks.
We centre our study on three key questions.

First we consider: \emph{Which models perform well on fine-grained tasks?}
To answer this, we evaluate models from four different model families trained with different amounts of pretraining data, as well as recent architectures that leverage frozen large language models (LLMs).
We observe that \textbf{modelling innovations have more impact than simply scaling image captions} from the Web.
Furthermore, explicitly modelling localisation can improve performance, but it is crucial \emph{how} it is done, and simply using localisation data is not enough.  

Our observations motivate our next question: 
\emph{How do data and losses impact fine-grained understanding?} We focus our study on the best performing model, \xvlm~\cite{x-vlm}, which learns to map specific objects and regions (not a full image) to a label (word or phrase describing the region).
We reformulate the \xvlm loss to better disentangle the contribution of data and losses, observing that more data does not improve performance unless paired with \textbf{losses designed to learn a mapping between regions and labels}.
Furthermore, the diversity of class labels is important for performance on coarse-grained retrieval, and region descriptions (as opposed to single word labels) are crucial for performance on fine-grained tasks.

Finally, it is unclear if all fine-grained skills are learned at the same time during training so we consider: \emph{How does fine-grained understanding evolve during training?}
Surprisingly, we find that while performance steadily improves on coarse-grained retrieval tasks through training, \textbf{performance fluctuates substantially on many fine-grained tasks}, with some skills, like counting, becoming increasingly \emph{worse}.
Additionally, performance across different fine-grained tasks that should test for similar skills are not always well correlated.

\textbf{Contributions}. In this work, we \textbf{1)} provide in-depth analyses of how data and modelling decisions impact performance on fine-grained tasks, and \textbf{2)} further  disentangle the gains given by data and pretraining losses on our best performing model (\xvlm).
Our results suggest that to make progress in fine-grained understanding, modelling innovations (\eg, through object-centric losses) as well as data quality and richness are more effective than scaling up Web data alone.
Finally, we \textbf{3)} shed light on VLMs' pretraining dynamics and suggest that future work should revisit pretraining strategies in order to consistently improve across several tasks.

\begin{table}[t!]
    \setlength{\tabcolsep}{2pt}
    \small
    \centering\resizebox{\linewidth}{!}{
\definecolor{LightCyan}{rgb}{0.96,0.96,0.96}

\hypersetup{citecolor=lightgray}
\begin{tabular}{l l rc l}
\toprule
\textbf{Benchmark} & \textbf{Task} & \textbf{Examples} & \textbf{Subtasks} & \textbf{Example Subtasks}  \\
\midrule
 \multicolumn{5}{c}{\emph{Fine-grained Tasks}} \\
 \midrule
 \rowcolor{LightCyan}
SVO-Probes  & Verb            & 48K & 3 & subject, verb, object \\
\rowcolor{LightCyan}
 & understanding            &  &  &  \\
VALSE       & \vl                 & 14K & 6 & existence, counting,  \\
 & grounding            &  &  &  spatial relations\\
 \rowcolor{LightCyan}
VSR         & Spatial             & 2K  & 7 & adjacency, directional, \\
\rowcolor{LightCyan}
 & reasoning           &  &  &  proximity relationships\\
Winoground  & Compositional      & 800 & 8 & pragmatics, object swap, \\
 & reasoning           &  &  &  relation swap\\
\midrule
\multicolumn{5}{c}{\emph{Coarse-grained Tasks}} \\
\midrule
\rowcolor{LightCyan}
COCO & Retrieval &  25K & 0  & N/A\\
Flickr30K & Retrieval &  5K & 0 & N/A \\
\bottomrule
\end{tabular}
    }
    \vspace{-2mm}    
    \caption{Overview of our benchmarks. For consistency, we report the number of examples as the number of positive image--text pairs in each evaluation dataset.}
    \label{tab:benchmarks}
    \vspace{-2mm}
\end{table}

\section{Benchmarks} \label{sec:benchmarks}

We describe the recent (English) benchmarks proposed to measure fine-grained \vl understanding in zero-shot setups.\footnote{We note that two more datasets require fine-grained skills to be solved and that they are not part of our analysis. ImageCoDe~\cite{imagecode} requires comparing a caption within a multi-image context, a setup not suitable for zero-shot evaluation of current single-image VLMs. \citet{vl_bow} propose the ARO benchmark to evaluate VLMs’ attribution, relation, and order understanding. However, the data had not been released as of the ACL deadline.}
See~\cref{tab:benchmarks} for an overview.

\begin{table*}[t!]
    \setlength{\tabcolsep}{3pt}
    \small
    \centering
    \resizebox{\linewidth}{!}{
        \begin{tabular}{l|ccc|ll|ccc}
\toprule
\multicolumn{1}{l}{\textbf{Model}} & \multicolumn{3}{c}{\textbf{Loss}} & \multicolumn{2}{c}{\textbf{Data}} & \multicolumn{3}{c}{\textbf{Downstream}} \\
 & CL & Text & Obj Det & \multicolumn{1}{c}{Unsupervised} & \multicolumn{1}{c|}{Supervised} & VQAv2 & NLVR2 & RefCOCO+ \\
\midrule
\albefb         & \checked & MLM & - & 4M: COCO+SBU+VG+CC$_\text{3M}$ & \multicolumn{1}{c|}{-}               & 74.7 & 80.5 & - \\
\albefl         & \checked & MLM & - & 14M: 4M + CC$_\text{12M}$ & \multicolumn{1}{c|}{-}                    & 76.0 & 83.1 & - \\
\midrule
\blipl          & \checked & LM & - & \capfiltb{}(14M)         & \multicolumn{1}{c|}{-}                      & 77.6 & 82.3 & - \\
\blipxl         & \checked & LM & - & \capfiltb{}(14M + LAION) & \multicolumn{1}{c|}{-}                      & 78.2 & 83.1 & - \\
\blipxlfilt     & \checked & LM & - & \capfiltl{}(14M + LAION) & \multicolumn{1}{c|}{-}                      & 78.3 & 82.2 & - \\
\blipvitxl      & \checked & LM & - & \capfiltl{}(14M + LAION) & \multicolumn{1}{c|}{-}                      & - & - & - \\
\midrule
\pevlpre        & \checked & MLM & MLM & 14M & RefCOCO\{,+,g\}+F30KE+GQA+VCR+VG                             & - & - & 74.5 \\
\midrule
\xvlmb          & \checked & MLM & Regress & 4M  & COCO + VG                                                & 78.1 & 84.2 & 71.0 \\
\xvlml          & \checked & MLM & Regress & 14M & COCO + VG + Objects365 + OpenImages                      & 78.4 & 84.4 & 76.9 \\
\bottomrule
\end{tabular}

    }
    \vspace{-2mm}
    \caption{Overview of core evaluated models. All the models use contrastive learning (CL), cross-attention and a (masked) language modelling objective. Fine-grained models also predict object locations from supervised data.}\label{tab:models}
    \vspace{-2mm}
\end{table*}

\textbf{SVO-Probes}~\cite{svo_probes} focuses on verb understanding: it tests whether a model can identify if an image matches a sentence, and includes negative images which differ on a specific part of speech (Subject, Verb, and Object).
The dataset consists of 421 verbs and over 48K image–sentence pairs.\footnote{Only 30{,}578 pairs were available as of Nov 2022.}
The authors show that their baselines fail more in situations requiring verb understanding than other parts of speech.

\textbf{VALSE}~\cite{valse} consists of six tasks that cover basic linguistic phenomena, such as plurality, actions and coreference.
For each task, given a visual input, a model is asked to distinguish real captions from foils~\cite{foil}, where a foil is constructed from a caption by altering a word or phrase that realises a specific linguistic phenomenon (\eg, semantic number of nouns).
The authors show that VLMs can identify objects in images, but struggle to ground their interdependence with specific linguistic indicators. %

\textbf{VSR}~\cite{vsr} tests for 65 types of visual spatial relationships (\eg, \texttt{under}, \texttt{in front of}) grouped into seven categories (\eg, adjacency, orientation).
Each sample consists of an image--sentence pair; a model needs to predict whether the sentence correctly describes the spatial relation between two objects in the image.
We evaluate models in a zero-shot setup on the `random' split.\footnote{Note that VSR has recently been updated, but we expect the findings from our experiments to hold on the revised splits.}

\textbf{Winoground}~\cite{winoground} is an expert-curated benchmark aiming to test models' compositional reasoning. 
Given two images and two captions, the goal is to match them correctly; wherein both captions contain the same set of words, but in a different order.
The authors define three scores: Text (whether a model can match the correct caption for a given image), Image (vice versa), and Group (whether a model can match each pair).
Several competitive VLMs have been shown to often perform close to or below random chance.

We also report zero-shot performance on coarse-grained retrieval in \textbf{Flickr30K}~\cite{flickr30k} and \textbf{COCO}~\cite{coco} in our analysis.
\section{Evaluated Models} \label{sec:eval_models}

Recent work has shown that two components are crucial ingredients of strong coarse-grained VLMs~\citep[\eg,][]{albef,alayrac2022flamingo,pali}: \textbf{1)} a contrastive objective that aligns vision and language modalities, and \textbf{2)} a cross-attention mechanism that fuses the two modalities. 
As we are interested in high performance on both fine- and coarse-grained tasks, to select models for our study, we surveyed recent work that uses these building blocks,\footnote{By studying models with well-established modules, we expect our findings to be more informative for future work.} but also incorporates new losses or data that can potentially improve fine-grained \vl understanding. 
We find that many recent models build on \albef~\cite{flava,tcl,mixgen} (which we also study as a coarse-grained baseline).

Other than strong performance on coarse-grained and downstream tasks, we also considered: \textbf{1)} the possibility to study the role of new modelling innovations and data for fine-grained skills, and \textbf{2)} the availability of open-source code and pretrained weights.
This resulted in four models briefly described next (more details in~\cref{app:models}).
\cref{tab:models} codifies the main differences in pretraining objectives and data used by these models.
Recall that previous work does not evaluate these models on fine-grained benchmarks.

\textbf{ALBEF}~\cite{albef}, with strong downstream performance, matches all our criteria and serves as a coarse-grained baseline.  \albef is a dual-stream encoder~\cite{unmasked} that first encodes images and captions independently, and then fuses them with cross-modal attention.

\textbf{BLIP}~\cite{blip} uses an autoregressive language model (LM), and employs a dataset bootstrapping technique (CapFilt) to generate synthetic captions and to remove noisy pairs from large-scale Web data.
BLIP outperforms \albef on most coarse-grained downstream tasks; thus, we study BLIP as another coarse-grained baseline to test if its generative LM and data contributions also lead to better fine-grained understanding.

\textbf{PEVL}~\cite{pevl} is a fine-grained model building on \albef, but leverages more supervised datasets such as referring expressions, captions with visual coreferences, object detection and region descriptions data, etc. (see \cref{tab:models}).
Unlike \albef, \pevl is explicitly trained to learn fine-grained representations of entities by predicting their coordinates in a unified masked language modelling framework \citep[similar to Pix2Seq,][]{pix2seq_mtl}: bounding box coordinates corresponding to a given entity are added in the caption as ``\texttt{A cat < 10 73 206 175 > is napping.}''

\textbf{X-VLM}~\cite{x-vlm} is our second fine-grained model that enhances \albef by adding both new losses and additional supervised data.
In contrast to \pevl, \xvlm models visual position through an additional bounding box prediction head that regresses the object's bounding box (bbox) coordinates.
The authors use both object detection labels and region descriptions to learn coarse- and fine-grained alignments (we provide an in-depth analysis of this model in \cref{sec:x-vlm}).

\medbreak
\noindent
We remark that \pevl and \xvlm were the only open-source fine-grained VLMs at the time of our evaluation, and both of them build on top of \albef.
In addition to these core models, we also evaluate a dual-encoder network (\clip; \citealt{clip}) as well as recent architectures that rely on frozen, autoregressive (L)LMs: \clipcap~\citep{clipcap}, \flamingo~\citep{alayrac2022flamingo} and \bliptwo~\citep{blip2}.
As these models perform generally worse than our best fine-grained model, \xvlm, and differ significantly from it, we do not discuss their performance further. For more details, we refer the reader to Tables\cref{tab:overall_baselines_app,tab:winoground_vis,tab:winoground_ling,tab:svo,tab:valse,tab:vsr} in~\cref{app:res_subtasks}.
\section{Which Fine-grained Models Perform Well on Fine-grained Tasks?} \label{sec4}

We compare two strong VLMs (\albef and \blip) with two models with explicit object modelling (\ie, fine-grained; \xvlm and \pevl).
We evaluate on fine-grained tasks (see \cref{tab:overall_baselines}) to determine if recent object-centric models improve on tasks designed to measure fine-grained skills---an evaluation missing from previous work.  
We also include results on \clip and \bliptwo in \cref{tab:overall_baselines} to highlight how well fine-grained models perform, even though pretrained with less data and having fewer parameters (as shown in \cref{tab:overall_baselines_app} in \cref{app:res_subtasks}).

\paragraph{Experimental setup.}
All our fine-grained benchmarks only require models to predict a matching score for a given image--text pair, a common task that current \vl models---including all of our evaluated models---are pretrained to solve.
On VSR, a model's prediction is correct if the matching score is greater/lower than 50\% for a true/false label.
On the other benchmarks, a model's prediction is correct if the score for the positive image--text pair is higher than the score of the negative pair(s).\footnote{We evaluate SVO-Probes using \emph{pairwise ranking accuracy} to benchmark models without a binary classification head (we note that \citealt{svo_probes} used accuracy).} 
We evaluate the public models released by the authors on GCP.\footnote{\url{https://cloud.google.com/}.} 
Code to reproduce our analysis is online.\footnote{\url{https://github.com/e-bug/fine-grained-evals}.}

\begin{table}[t!]
    \setlength{\tabcolsep}{2pt}
    \small
    \centering
    \resizebox{\linewidth}{!}{
        \hypersetup{citecolor=lightgray}
\begin{tabular}{llcccccc}
\toprule
& \multicolumn{1}{l}{\textbf{Model}} & \multicolumn{1}{c}{\textbf{SVO}} & \multicolumn{1}{c}{\textbf{VALSE}} & \multicolumn{1}{c}{\textbf{VSR}} & \multicolumn{3}{c}{\textbf{Winoground}}  \\
& & Avg. & Avg. & Test Avg. & Text & Image & Group \\
\midrule
\multicolumn{1}{r|}{} & \gray{Random}         & \gray{50.0} & \gray{50.0} & \gray{50.0} & \gray{25.0} & \gray{25.0} & \gray{12.5}  \\
\midrule
\multicolumn{1}{r|}{} & \gray{\clip{}$_\text{400M}$}         & \gray{81.6} & \gray{64.0} & \gray{N/A} & \gray{30.7} & \gray{10.5} & \gray{~~8.0}  \\
\multicolumn{1}{r|}{} & \gray{\bliptwo{}$_\text{129M}$}         & \gray{86.5} & \gray{74.0} & \gray{61.5} & \gray{43.0} & \gray{22.0} & \gray{18.2}  \\
\midrule
\multicolumn{1}{r|}{1} & \albefb         & 87.6 & 69.1 & 57.3 & 29.2 & 15.5 & 11.0  \\
\multicolumn{1}{r|}{2} & \xvlmb{}$^\sharp$       & \underline{88.9} & \underline{72.4} & \underline{63.0} & \underline{44.0} & \textbf{26.7} & \textbf{21.5} \\
\midrule
\multicolumn{1}{r|}{3} & \albefl         & 88.6 & 69.4 & 58.3 & 32.5 & 16.2 & 12.7 \\
\multicolumn{1}{r|}{4} & \blipl          & 48.7 & 67.8 & 49.7 & 36.5 & 18.5 & 14.5 \\
\multicolumn{1}{r|}{5} & \pevll{}$^\sharp$       & 86.2 & 68.9 & 57.5 & 33.2 & 15.7 & 12.2  \\
\multicolumn{1}{r|}{8} & \xvlml{}$^\sharp$        & \textbf{90.0} & \textbf{74.5} & \textbf{64.3} & \textbf{46.7} & \underline{24.5} & \underline{21.2} \\
\midrule
\multicolumn{1}{r|}{9} & \blipxl         & \underline{51.4} & 68.8 & 46.9 & \underline{35.5} & 15.0 & 11.7 \\
\multicolumn{1}{r|}{10} & \blipxlfilt     & 51.2 & 68.2 & 48.7 & 34.7 & \underline{15.2} & \underline{12.2} \\
\multicolumn{1}{r|}{11} & \blipvitxl      & 50.8 & \underline{70.3} & \underline{50.3} & 34.7 & 14.5 & \underline{12.2} \\
\bottomrule
\end{tabular}

    }
    \vspace{-2mm}
    \caption{Overall performance of core evaluated models on fine-grained benchmarks; the highest values for a given data size and the overall best values are marked with underline and bold, respectively. $^\sharp$ marks fine-grained models.
    For a breakdown of task performance and full comparison with prior arts, see~\cref{app:res_subtasks}.}
    \label{tab:overall_baselines}
    \vspace{-4mm}
\end{table}

\paragraph{ALBEF vs. BLIP.}
We first compare our two coarse-grained baselines.
A key difference between \albef and \blip is that the former is trained with masked language modelling (MLM), while the latter uses autoregressive language modelling (LM) for text; with \blip outperforming \albef on downstream tasks when pretrained on the same 14M images.
Performing the same comparison on fine-grained benchmarks, we find that \albefl outperforms \blipl on all tasks (largely on SVO-Probes and VSR) except on Winoground.
Likewise, \cref{tab:overall_baselines_app} (\cref{app:res_subtasks}) shows that other visual-conditional LMs, such as \clipcap models, also struggle with fine-grained understanding.
This might be due to the fact that our evaluation relies on image--text alignments and does not test for generation, where the LM objective is often preferred. 
Given these results and the fact that \albef is more similar to our fine-grained models, we compare against \albef in most of our discussion.

\paragraph{Effectively modelling object positions improves fine-grained understanding.}
Overall, we find that \xvlm consistently outperforms all other evaluated approaches (see \Table{tab:overall_baselines}). This trend holds in both the 4M and 16M pretraining setups. When trained on the same 4M images as the \albef baseline, \xvlm with explicit object modelling, notably improves over all benchmarks (gaining 1.3pp on SVO-Probes, 3.3pp on VALSE, 5.7pp on VSR, and 14.8/11.2/11.5pp on Winoground).
Importantly, \xvlmb also outperforms \albefl (trained on 10M more data points).
This result shows the importance of explicit object modelling for a range of fine-grained tasks, including ones that are dissimilar to the supervised localisation task (\eg, verb understanding).

\xvlml, which adds CC$_\text{12M}$ as well as object detection data from OpenImages and Objects365 to \xvlmb's data, achieves even higher overall gains in most fine-grained benchmarks.
On VALSE, it closes the gap with a larger model trained on supervised data from many downstream tasks (12-in-1;~\citealt{12in1}), and on VSR it achieves similar accuracy to LXMERT~\citep{lxmert} fine-tuned on 50\% of VSR training data (67.9pp).
Moreover, on Winoground, \xvlmb significantly outperforms previous coarse-grained models, including a large-scale dual-encoder \citep[\clip, Group score of 8.0;][]{clip} and a strong, larger cross-modal Transformer \citep[UNITER$_\text{Large}$, Group score of 10.5;][]{uniter}, as shown in \cref{tab:overall_baselines_app} in \cref{app:res_subtasks}.

\paragraph{Not all object modelling improves fine-grained understanding.}
Like \xvlm, \pevl also models visual locations of objects.
However, it does so by expecting (masked) bbox locations as part of its input caption.
Surprisingly, \pevll performs much worse than \xvlml on all tasks; in fact, it performs on par with the \albefl baseline, despite being originally initialised with its checkpoint and further tuned to model visual object locations.\footnote{We evaluate three different models released by the authors, which differ in their pretraining and fine-tuning data. All the variants perform similarly, and as a result, we only report \pevlpre, which underwent a second-stage pretraining on multiple supervised tasks (\cref{app:res_subtasks} lists all the models).}
We conjecture that modelling objects as input prompts is less beneficial than directly predicting object locations with a classification head (\xvlm), as the former does not directly influence the object's representations in the text modality.

\paragraph{Modelling objects has more impact than increasing data.}
In \cref{tab:overall_baselines}, we observe that, not surprisingly, increasing data for a given family (\eg, \albefb to \albefl) results in improved performance on most benchmarks.
However, interestingly, the \emph{fine-grained} \xvlmb, trained on 4M data points, outperforms all \blipxl variants---a coarse-grained model trained on $129$M data points (compare row 2 with rows 9--11). 
Similarly, while increasing the data from 4M to 14M results in improvements across most tasks for the coarse-grained \albefl, these performance gaps are smaller than what we gain from modelling objects on top of \albefb.
That is, the average performance gap between \albefb and \xvlmb is bigger (+5.2pp) than that observed when increasing data from \albefb to \albefl (+1.0pp).
This result highlights that simply scaling data, without modelling innovations, might not be enough for notable improvements on fine-grained tasks.

\begin{figure*}[t!]
    \centering
    \includegraphics[width=\textwidth, trim={0cm 8.4cm 0.3cm 0cm}, clip]{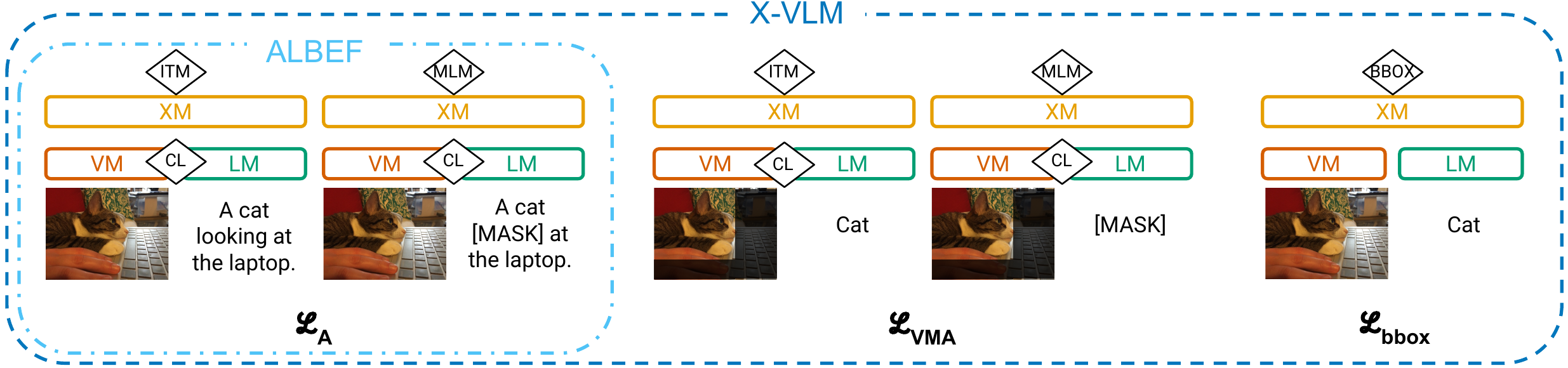}
    \vspace{-7mm}
    \caption{Overview of the pretraining objectives effectively used by \albef and \xvlm.}\label{fig:xvlm_losses}
    \vspace{-3mm}
\end{figure*}

We also find that scaling data can \emph{hurt} performance on some benchmarks.
For example, on Winoground Image and Group scores, \xvlml and \blipvitxl perform worse than their corresponding models trained on less data, \xvlmb and \blipl, respectively.\footnote{While \blipxl performs worse than \blipl on a few benchmarks, this might be because the data size is significantly increased without scaling the model size. Thus, we compare against \blipvitxl, which uses a larger image encoder.}
Looking at performance by subtasks, we find that scaling Web data leads to worse performance on several of them, such as Image scores in most Winoground tasks, and VALSE's \texttt{existence}, \texttt{counting adversarial} and \texttt{coreference} for \blipvitxl (more details in~\cref{app:res_subtasks}).
We conjecture that pretraining on noisy Web data---where the language in an image--text pair does not always faithfully describe the image---might diminish the fine-grained alignments learned from smaller, cleaner datasets (\citealt{hendricks2021decoupling} report similar trends on coarse-grained tasks).

\paragraph{Takeaways.}
We observe that modelling object positions in images provides a strong signal for fine-grained understanding; but, \emph{how} we model this information is crucial: simply pretraining a model with bbox positions in input does not lead to better off-the-shelf representations.
We also see bigger gains on fine-grained tasks when modelling objects compared to scaling the pretraining data.

\section{Data \& Losses for Fine-grained Tasks} \label{sec:x-vlm}

Recent fine-grained models build on coarse-grained ones by introducing additional training data (\eg, object detection data in \xvlm and \pevl) and new losses (\eg, bounding box regression loss in \xvlm).
We study how data and losses influence fine-grained understanding, focusing on \xvlm as it outperforms other models on fine-grained benchmarks.
While \citet{x-vlm} perform ablations to show the importance of their new objective function, they do not study the impact of data and losses independently; moreover, they do not evaluate on find-grained benchmarks.
We start with a description of \xvlm, emphasising details in its pretraining procedure, that we reveal to have significant impact on the final performance. 

\subsection{What are X-VLM Data and Losses?}

The \xvlm architecture consists of the same modules as \albef: a vision, a text, and a cross-modal Transformer~\citep{transformer} encoder (see \cref{app:models} for details).
Given an image--text pair, \albef performs two forward passes (as shown in \cref{fig:xvlm_losses}): first, the model computes a contrastive learning loss ($\mathcal{L}_\text{CL}$) and an image--text matching loss ($\mathcal{L}_\text{ITM}$).  In a second pass, it masks text inputs to compute a visually-grounded masked language modelling loss, $\mathcal{L}_\text{MLM}$.
After the two forward passes, \albef is trained with $\mathcal{L}_\text{A} = \mathcal{L}_\text{CL}+\mathcal{L}_\text{ITM}+\mathcal{L}_\text{MLM}$.

\paragraph{Data.}
While \albef is only pretrained on image--caption data, \xvlm additionally pretrains on object and region detection data.  Object detection data consists of an object or attribute--object label (\eg, ``dog'' or ``brown dog''), an image, and a bounding box; region detection data consists of a short phrase (\eg, ``a cute brown dog''), an image, and a bounding box.
Other multimodal Transformer models have used detection data \citep{hendricks2021decoupling,oscar,unmasked,vinvl}, but usually the bounding boxes are discarded, and objects or region descriptions are paired with the \emph{entire} image.
In contrast, a close examination of the \xvlm codebase\footnote{\url{https://github.com/zengyan-97/X-VLM}.} reveals that \xvlm effectively makes use of bounding boxes.

\paragraph{BBOX loss.} 
To take advantage of additional bounding box (bbox) data, \xvlm introduces an objective, $\mathcal{L}_\text{bbox}$, which learns to regress to object locations from object detection and region description data (see \cref{fig:xvlm_losses} for an overview).

\begin{table*}[t!]
    \setlength{\tabcolsep}{3.0pt}
    \small
    \centering
    \resizebox{\linewidth}{!}{
        \begin{tabular}{lcccc|ccc|ccc|cccc}
\toprule
& \multicolumn{4}{c}{\textbf{Data}} & \multicolumn{3}{c|}{\textbf{Loss}} & \multicolumn{1}{c}{\textbf{SVO-Probes}} & \multicolumn{1}{c}{\textbf{VALSE}} & \multicolumn{1}{c|}{\textbf{VSR Random}} & \multicolumn{2}{c}{\textbf{Flickr30K}} & \multicolumn{2}{c}{\textbf{COCO}} \\
& \multicolumn{1}{|c}{$\mathcal{D}_\text{A}$} & COCO$_\text{OD}$ & VG$_\text{OD}$ & VG$_\text{RD}$ & $\mathcal{L}_\text{A}$ & $\mathcal{L}_\text{VMA}$ & $\mathcal{L}_\text{bbox}$ & Avg. & Avg. & Test Avg. & TR@1    & IR@1   & TR@1   & IR@1  \\
\midrule
\multicolumn{1}{r|}{0} & \checked & & & & \checked & & &                                                85.9 & 68.7 & 59.3 & 76.3 & 59.8 & 60.9 & 45.7 \\
\midrule
\multicolumn{1}{r|}{1} & \checked & \checked & & & \textcolor{lightgray}{\checked} & \textcolor{lightgray}{\checked} & \textcolor{lightgray}{\checked} &                     85.9 & 69.1 & 58.6 & 72.8 & 59.5 & 60.8 & 46.1 \\
\multicolumn{1}{r|}{2} & \checked & & \checked & & \textcolor{lightgray}{\checked} & \textcolor{lightgray}{\checked} & \textcolor{lightgray}{\checked} &                     86.0 & 68.6 & 59.7 & 77.1 & 62.7 & 63.3 & 47.5 \\
\multicolumn{1}{r|}{3} & \checked & & & \checked & \textcolor{lightgray}{\checked} & \textcolor{lightgray}{\checked} & \textcolor{lightgray}{\checked} &                     86.6 & 70.3 & 61.1 & 79.4 & 62.3 & \textbf{64.8} & \textbf{49.1} \\
\multicolumn{1}{r|}{4} & \checked & \checked & \checked & & \textcolor{lightgray}{\checked} & \textcolor{lightgray}{\checked} & \textcolor{lightgray}{\checked} &            85.6 & 67.5 & 60.7 & 77.2 & 60.7 & 63.3 & 47.3 \\
\multicolumn{1}{r|}{5} & \checked & \checked & & \checked & \textcolor{lightgray}{\checked} & \textcolor{lightgray}{\checked} & \textcolor{lightgray}{\checked} &            86.5 & 67.6 & 60.1 & 77.2 & 61.4 & 62.9 & 47.6 \\
\multicolumn{1}{r|}{6} & \checked & & \checked & \checked & \textcolor{lightgray}{\checked} & \textcolor{lightgray}{\checked} & \textcolor{lightgray}{\checked} &            \textbf{86.9} & \textbf{71.1} & \textbf{62.5} & \textbf{79.7} & \textbf{63.4} & 64.4 & \textbf{49.1} \\
\midrule
\multicolumn{1}{r|}{7} & \textcolor{lightgray}{\checked} & \textcolor{lightgray}{\checked} & \textcolor{lightgray}{\checked} & \textcolor{lightgray}{\checked} & \checked & & &                     85.9 & 69.3 & 58.2 & 75.5 & 58.9 & 61.9 & 45.8 \\
\multicolumn{1}{r|}{8} & \textcolor{lightgray}{\checked} & \textcolor{lightgray}{\checked} & \textcolor{lightgray}{\checked} & \textcolor{lightgray}{\checked} & \checked & \checked & &            86.5 & 69.1 & 59.0 & 77.5 & 62.3 & 63.0 & 47.6 \\
\multicolumn{1}{r|}{9} & \textcolor{lightgray}{\checked} & \textcolor{lightgray}{\checked} & \textcolor{lightgray}{\checked} & \textcolor{lightgray}{\checked} & \checked & & \checked &            86.0 & 67.9 & 60.5 & 78.0 & 60.5 & 62.1 & 47.6 \\
\midrule
\multicolumn{1}{r|}{10} & \checked & \checked & \checked & \checked & \checked & \checked & \checked &  \textbf{86.9} & 69.8 & 61.9 & 78.3 & 63.0 & 64.6 & 48.6 \\
\bottomrule
\end{tabular}
    }
    \vspace{-2mm}
    \caption{Overall performance of \xvlm ablations pretrained on the \emph{exact same} data. Rows 0 and 10 are our re-implementation of \albef and \xvlm, respectively. Rows 3, 4, and 8 correspond to ``w/o object,'' ``w/o region,'' and ``w/o bbox'' ablations in \citet{x-vlm}.
    We find that $\mathcal{L}_\text{VMA}$ is crucial towards \xvlm's performance, and that \vgrd yields richer signal for both coarse- and fine-grained tasks.}
    \label{tab:overall_ablations}
    \vspace{-3mm}
\end{table*}

\paragraph{VMA loss.}  
The \xvlm paper presents two losses, $\mathcal{L}_\text{A}$ and $\mathcal{L}_\text{bbox}$.
However, $\mathcal{L}_\text{A}$ operates over two input types: image--text pairs from captioning data and image--text--bbox triplets from object detection data.
Thus, it is hard disentangle the impact of the data and the losses on performance.
We reformulate $\mathcal{L}_\text{A}$ into two losses,\footnote{Our reformulation is equivalent to \xvlm, but it allows us to disentangle the impact of data and losses on performance.} operating over: (a) image--text pairs, $\mathcal{L}_\text{A}$, as in \albef; or (b) image--text--bbox pairs, that we denote \emph{visually masked} \albef loss, $\mathcal{L}_\text{VMA}$.
For $\mathcal{L}_\text{VMA}$, the visual and cross-modal encoders only attend to the image patches that correspond to the object bbox coordinates via an attention mask (see \cref{fig:xvlm_losses}).
This results in an object-centric visual view for grounding the text label through the pretraining objectives.
To compute this loss, in addition to the three forward passes described so far (CL and ITM, MLM, and BBOX losses), \xvlm performs two more passes: one where image patches outside a bounding box region are masked out to compute the \emph{visually masked} CL and ITM loss, and another where text is additionally masked for the \emph{visually masked} MLM loss.
\cref{sec:xvlm-ablate} shows both the data and pretraining techniques are key to the final model performance.

\subsection{Experimental Setup}
We re-implement ALBEF and X-VLM in JAX to ensure full control of modelling, data, and initialisation decisions.\footnote{To verify our implementation, we compare an \albef model trained in our codebase with one trained in the original codebase, obtaining an absolute difference below 1pp in Recall@1 on zero-shot Flickr30K and COCO retrieval tasks.}
We initialise both models with a 224$\times$224 ViT-B/16 visual encoder~\cite{vit_augreg}, and BERT$_\text{BASE}$~\cite{bert} weights in the text and cross-modal layers.
Similar to~\citet{unmasked}, we pretrain our models on the \emph{exact same} 4M and 14M datasets used by the authors~(\cref{tab:models}), but note that only 1.8M and 11.2M data points were available for CC$_\text{3M}$ and CC$_\text{12M}$, respectively.
For object detection data, we use the COCO and VG annotations released by the \xvlm authors.
Following \citet{x-vlm}, we pretrain our models for 200K steps using the official hyperparameters (see \cref{app:exp_setup} for more details).

\subsection{Results} \label{sec:xvlm-ablate}

\cref{tab:overall_ablations} shows the overall zero-shot performance of our ablations on three fine-grained benchmarks and two coarse-grained retrieval tasks.
Row 0 is our \albef re-implementation, while row 10 corresponds to our \xvlm pretrained following the implementation of \citet{x-vlm}.
Our controlled study allows us to quantify how each technique (losses, data, implementation details) in \xvlm contributes towards fine-grained understanding.

\paragraph{Data ablation.}
We first investigate the role of supervised detection data used to learn fine-grained relationships in \xvlm by pretraining the model, using its standard training objectives, and adding different data sources (rows 1--6).

Looking at rows 1--3, we find that region descriptions from VG (\vgrd) are the most useful, single-source signal for the model, resulting in improvements in both fine- and coarse-grained tasks. 
This variant is either close to or surpasses the final \xvlm variant (row 10) in all the tasks.
We attribute this success to both its size (3.7M data points) and language format, wherein noun phrases, rather than simple labels, describe a given entity.
In addition, object detection data from VG (\vgod) leads to similar fine-grained results as \cocod, but significantly better zero-shot retrieval performance.
\vgod is not only larger than \cocod, but also includes a more diverse set of classes.\footnote{\cocod and \vgod have 80 and 50k labels respectively.} 

\begin{figure}[t!]
    \centering
    \includegraphics[width=0.48\textwidth, trim={0cm 0.0cm 0cm 0cm}, clip]{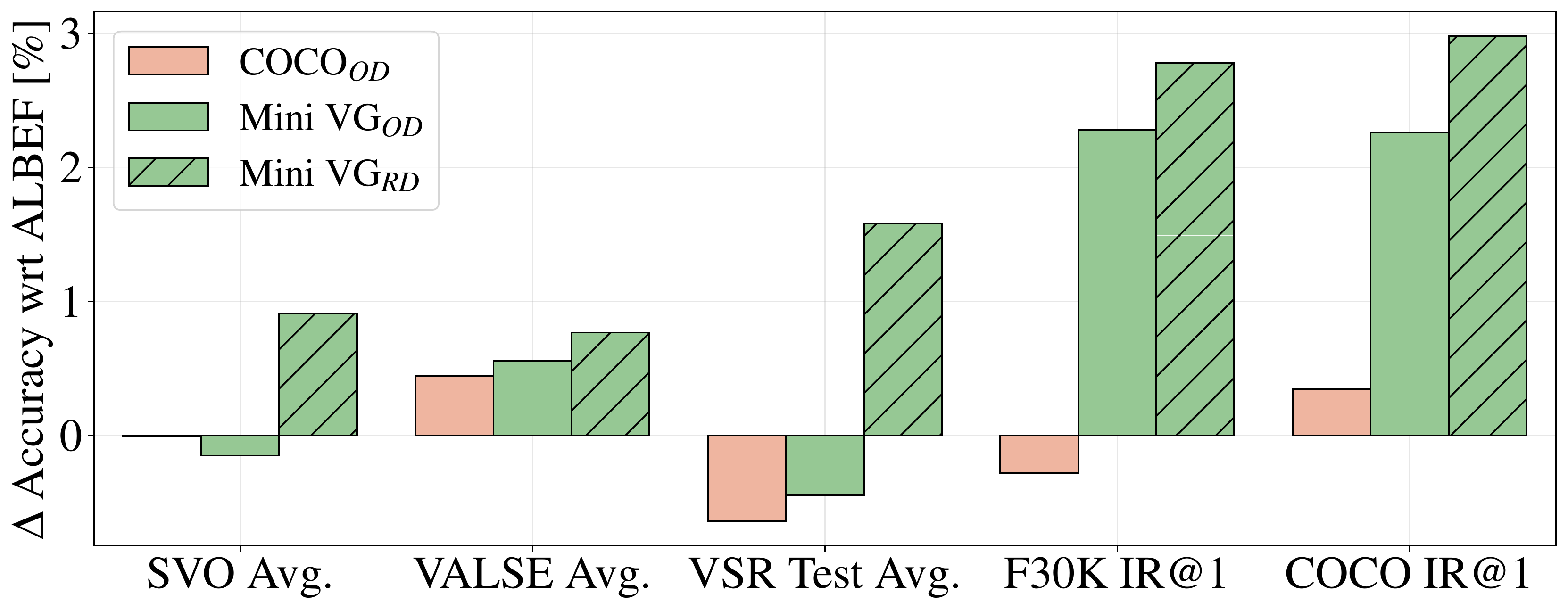}
    \vspace{-6mm}
    \caption{Performance on our benchmarks of \xvlm wrt \albef in a controlled setup with a single supervised dataset having the same number of images and annotations. Region descriptions give the highest gains.}\label{fig:mini_evals}
    \vspace{-3mm}
\end{figure}

We hypothesise that a \emph{large number of classes} (as in \vgod) is important for coarse-grained retrieval tasks, and \emph{more descriptive phrases} of \vgrd (rather than single labels) significantly impact fine-grained tasks.
To verify this, we disentangle the effect of data size and type: specifically, we re-train rows 2--3 on a subset of VG with the same number of images and annotations as in \cocod.
\cref{fig:mini_evals} confirms our hypothesis: even when controlled for size, \vgrd leads to notably better performance than \cocod.
On coarse-grained datasets, \vgod largely outperforms \cocod.

Looking at multi-source supervised data (rows 4--6), 
our best performing model combines \vgod and \vgrd data (row 6) and, surprisingly, adding \cocod does not boost performance.

\paragraph{Loss ablation.}
We investigate the role of the two objectives used during supervised pretraining of \xvlm (rows 7--9).
We see that training an \albef model on object detection data as-is (row 7) results in similar performance as pretraining it on standard image--caption data.
That is, \emph{just adding more data is not enough}; additional supervision in the form of the \xvlm pretraining objectives is crucial.

Compared to \lbbox (row 9), our reformulation makes it clear that \lvma (row 8) leads, on average, to both higher fine-grained accuracy and higher recall on retrieval tasks.
One potential explanation is that the visually masked forward pass directly influences the representation learned by the contrastive loss, as well as the cross-modal representations.  
In contrast, the regression loss only occurs after cross-modal interaction, suggesting that better alignment is important in both contrastive and cross-modal features.
Finally, \xvlm achieves its best performance when combining \lvma and \lbbox.

\begin{figure*}[t!]
    \centering
    \includegraphics[width=\textwidth, trim={0cm 0cm 0cm 0cm}, clip]{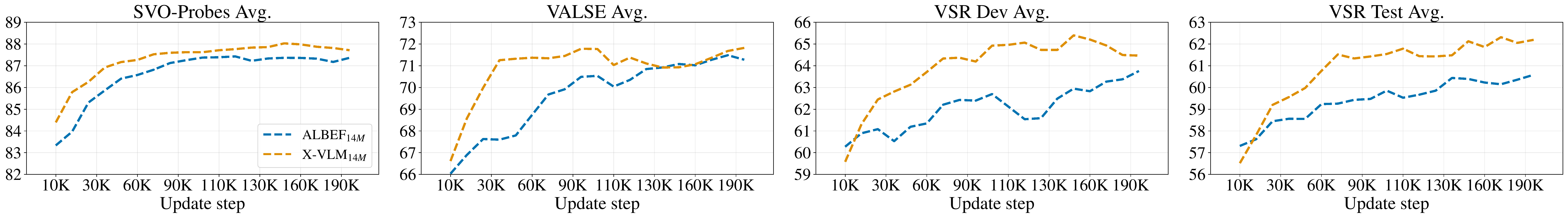}
    \includegraphics[width=\textwidth, trim={0cm 0cm 0cm 0cm}, clip]{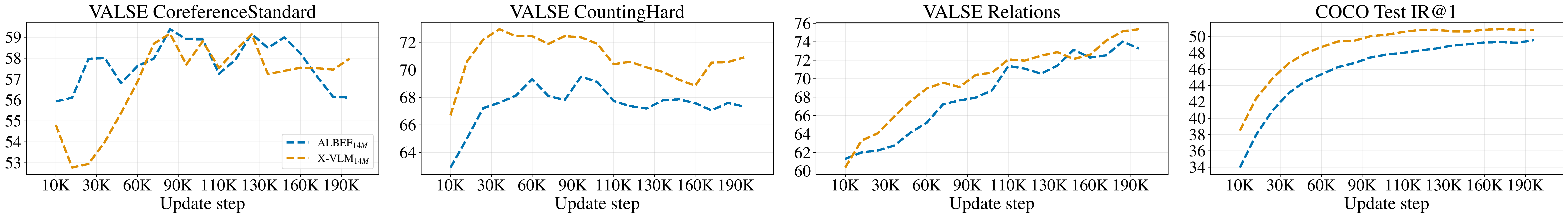}
    \vspace{-6mm}
    \caption{Training dynamics of our \albefl and \xvlmm models. Models' overall performance converges at different rates on different fine-grained benchmarks (top row). Performance on specific skills varies drastically, with some skills that deteriorate after an initial peak (bottom row). Retrieval performance (bottom right) does not capture this diversity in dynamics. Values smoothed with 0.6 factor for better visualisation. Full results in~\cref{app:res_dynamics}.}
    \label{fig:highlights_dynamics}
    \vspace{-3mm}
\end{figure*}

\paragraph{Takeaways.}
Our reformulation of \xvlm allows us to conduct a careful analysis in a controlled setup on how data and losses influence \xvlm performance. 
We show that more data does not improve performance unless paired with additional supervisory signal, in the form of either the visually masked ALBEF loss or bbox regression.
Given our observations and the fact that, as seen in \cref{sec4} and \cref{app:res_subtasks}, \xvlm largely outperforms the large-scale \clip{} and \bliptwo models on fine-grained tasks such as VALSE and Winoground, we believe that a promising direction in fine-grained understanding will require careful model and loss design with rich data sources like VG, not just scaling up with (potentially) noisy data.

\section{Dynamics of Fine-grained Tasks} \label{sec:dynamics}

We now analyse the dynamics of fine-grained skills for our models to investigate (i) when and whether they are acquired, and (ii) how they relate to one another, especially when they aim at measuring similar capabilities.
For example, does action understanding in VALSE correlate with verb understanding in SVO-Probes?
Are there skills that vastly differ from each other that they would require different modelling contributions (\eg, counting)?

\paragraph{Experimental setup.}
We evaluate checkpoints (every 10K steps) from pretraining our \albef and \xvlm re-implementations with 4M and 14M data points.
We focus on 14M results as we see similar patterns with 4M (see~\cref{app:res_dynamics}).
When evaluating correlation patterns, we report both Pearson and Spearman correlation coefficients.

\paragraph{Different skills, different patterns.}
\cref{fig:highlights_dynamics} (top) shows how the average model performance 
evolves during pretraining for the four benchmarks. Interestingly, the performance on these benchmarks converges at different rates: both \albef and \xvlm models easily improve on SVO-Probes. Moreover, we observe that modelling objects (à la \xvlm) leads not only to better fine-grained understanding after 200K steps (Tables~\cref{tab:overall_baselines,tab:overall_ablations}), but also to remarkably quicker learning rates.

\cref{fig:highlights_dynamics} (bottom) shows performance on indicative VALSE tasks, as well as on coarse-grained image retrieval on COCO.
While some skills, such as \texttt{spatial relations} understanding, are learned progressively during pretraining, others, such as \texttt{counting}, \emph{degrade} after a first, short learning phase.
Finally, other skills, such as \texttt{coreference resolution}, \emph{oscillate} significantly throughout pretraining, showing how models can not properly acquire them.
This is in contrast to the coarse-grained COCO retrieval task for which the performance steadily increases over time.
We conclude that it is particularly important to examine the training dynamics of fine-grained tasks, and that a single checkpoint might be inadequate for a number of skills.
Results on all tasks are provided in~\cref{app:res_dynamics}, including on Winoground for an \albefb that we pretrained on GCP using the original codebase.

\paragraph{Same skills, same patterns?}
We next investigate whether closely-related tasks in different benchmarks have high correlation throughout pretraining.
While we find that VALSE \texttt{action replacement} and SVO-Verb have a +55/67\% Pearson/Spearman correlation, there is -13/11\% correlation between VALSE \texttt{actant swap} and SVO-Subj.

Looking at VALSE \texttt{spatial relations}, we find high correlation (+75/65\%) with average VSR performance, and especially with relations such as \texttt{on top of}, \texttt{on}, \texttt{inside}, \texttt{by}, and \texttt{in}; mostly belonging to the `Topological' category in VSR.
On the other hand, we find almost no correlation with several `Directional' (\eg, \texttt{across from}) and `Orientation' (\eg, \texttt{parallel to}) relations, as well as with some `Topological' ones (\eg, \texttt{touching}); and even negative correlation (-40\% or less) with \texttt{alongside}, \texttt{below}, \texttt{toward}, \texttt{part of} and \texttt{near}.

Finally, surprisingly, VSR dev and test splits are \emph{not} positively correlated for all relations.
While average performance is highly correlated (+77/78\%), only a few relations have Pearson/Spearman coefficients larger than 30\% (\texttt{in}, \texttt{on}, \texttt{above}, \texttt{within}, and \texttt{consists of}).
On the other hand, \texttt{near}, \texttt{ahead of} and \texttt{adjacent to} are negatively correlated between dev and test sets, and most relations show very low correlations between the two sets.
As a result, improvement in a given relation type on the dev set, will likely not transfer at test time.

\paragraph{Takeaways.}
When tested on fine-grained benchmarks, we observe that, compared to \albef, \xvlm is more sample efficient as it achieves higher performance with fewer training steps.
Also, while some tasks steadily improve during pretraining, for others, the performance \emph{degrades} or \emph{fluctuates}.
Moreover, surprisingly, the performance of tasks measuring similar skills but from different benchmarks do \emph{not} always positively correlate.

\section{Discussion}

While recent pretrained VLMs achieve impressive performance on various downstream benchmarks (such as visual question answering and image retrieval), recent benchmarks have highlighted that they still struggle with tasks that require \emph{fine-grained} understanding---where a model needs to correctly align various aspects of an image to their corresponding language entities. 
Yet, it is still not known to which extent recent fine-grained VLMs~\citep[\eg,][]{x-vlm,pevl,loupe,fiber} fare on such benchmarks.
We address this gap by evaluating strong and fine-grained models on four benchmarks~\citep{svo_probes,valse,vsr,winoground}, and encourage future work to report zero-shot fine-grained performance on our selection of benchmarks, especially if models are not open-source.

Our work contributes to a growing thread of research devoted to understand what is learned by pretrained VLMs, such as studying cross-attention patterns~\citep{cao-etal-2020-behind}, cross-modal input ablations~\citep{v4l}, probing linguistic and visual structure~\citep{vl_structural,probing_vl,nikolaus-etal-2022-vision}, robustness to words order~\citep{akula-etal-2020-words,winoground}, and incorrectly fusing image and language modalities~\citep{why_wino}. 
Here, we show that object modelling through a prediction loss (as done in \xvlm) results in notable improvements across all benchmarks, outperforming models trained on much larger amounts of Web data.
Our analysis highlights that teaching VLMs concepts of objects (\eg, by masking irrelevant parts of the image) is crucial for effectively learning fine-grained skills. 
Though our models rely on supervised data to learn better localisation, we hope our findings can encourage researchers to design better loss functions for image--text mapping from unsupervised, Web-scale data as well.

Finally, our results also highlight the challenges of evaluating fine-grained understanding: the recent benchmarks capture a variety of subtasks (from counting to relation understanding); to perform well on these subtasks, a model requires different skills.
Indeed, we observe that, during training, model performance does not always increase for all subtasks, and in particular, fluctuates a lot for counting, coreference resolution, and various spatial relations.
An important future direction is designing models that perform well on a larger range of these subtasks, where improving on one subtask does not degrade performance on the rest.
It is unclear why benchmarks do not always correlate; possible reasons include the data itself (images selected for analysis, annotator instructions), or that different competencies are required for different fine-grained tasks. We hope future work can explore this further, possibly by closely examining data in fine-grained benchmarks or expanding the models used in analysis beyond what we used here.

\section*{Limitations}

Our work focuses on assessing recent English VLMs on tasks which require fine-grained understanding.
Here, we outline limitations that we believe are important considerations for future work.

First, we only examined a limited number of models.
These include (i) strong coarse-grained models, such as \albef, \clip, \flamingo and \bliptwo, and (ii) two strong fine-grained models, \pevl and \xvlm, that build on \albef.
While we believe our selection of models is representative of strong components in pretrained VLMs (such as dual-encoder and cross-modal interactions), we could not easily evaluate different approaches towards fine-grained understanding \citep[\eg,][]{filip,loupe} as the corresponding models and code are not open-source.
We hence hope our study will motivate future work to report zero-shot performance on fine-grained benchmarks.

Second, we evaluate our models in a zero--shot setting using image--text matching.
Future work could consider how fine-grained understanding improves when fine-tuning for specific tasks.
As opposed to relying on image--text matching scores, alternative methods like input ablations, visualising attention or activations could also be used to gain an understanding of potential failure modes.

Third, though we note specific areas where model performance fluctuates a lot during pretraining, we look forward to future research that improves performance for various such areas, like existence and counting.

Finally, some datasets we use are quite small.
For example, Winoground only has 1{,}600 data points.
We hope that our analysis sheds light on the kinds of skills models struggle with and encourages more and larger datasets that test for these skills.
\section*{Ethics Statement}

All datasets used in this work have been previously published.
Multimodal datasets frequently include social biases \cite{meister2022gender}, and we expect the models trained on them to reflect the biases in these datasets.
Datasets also include images of people, and there is no mechanism for people to remove themselves from these datasets.

Multimodal models have many downstream uses.
Some examples of beneficial applications include: more advanced image and video retrieval, visual description systems to aid the visually impaired, and interfaces which allow users to more seamlessly interact with smart home devices.
Harmful applications might include surveillance, especially when imagery of people is being used without their consent, or fine-tuning a model to retrieve harmful content, such as pornographic material.

In this work, we aim to understand how models perform on fine-grained tasks which highlights current failure modes of our models.
We hope insights from our work can inspire (i) novel models which perform well on a broad set of fine-grained tasks, as well as (ii) more high quality data to stress test our models.
We hope our work also helps those who might use multimodal models in downstream applications better anticipate how well these models might perform on their tasks.

\section*{Acknowledgements}
The authors would like to thank the anonymous reviewers, Antoine Miech, Ravichandra Addanki, Wojciech Stokowiec, Chris Dyer and the DeepMind Language Team for feedback on this project.

\bibliography{anthology,custom}
\bibliographystyle{acl_natbib}

\clearpage
\appendix

\section{Experimental Setup} \label{app:exp_setup}

In this section, we provide further details on the experimental setups that we used for our studies.

\begin{table*}[t!]
    \setlength{\tabcolsep}{3.5pt}
    \small
    \centering
    \resizebox{\linewidth}{!}{
        \begin{tabular}{llc|lrrr}
\toprule
\multicolumn{3}{c}{\textbf{Model}} & \multicolumn{4}{c}{\textbf{Data}} \\
Name & ViT & Img Res & Datasets & \# Img & \# Cap & \# Ann \\
\midrule
\albefb         & DeiT-B/16 & 256$\times$256 & 4M: COCO+SBU+VG+CC$_\text{3M}$ & 4.0M & 5.1M & - \\
\albefl         & DeiT-B/16 & 256$\times$256 & 14M: 4M + CC$_\text{12M}$ & 14.1M & 15.2M & - \\
\midrule
\blipl          & ViT-B/16 & 224$\times$224 & \capfiltb{}(14M) & 14.1M & 15.2M & - \\
\blipxl         & ViT-B/16 & 224$\times$224 & \capfiltb{}(14M + LAION) & 129.1M & 130.2M & - \\
\blipxlfilt     & ViT-B/16 & 224$\times$224 & \capfiltl{}(14M + LAION) & 129.1M & 130.2M & -  \\
\blipvitxl      & ViT-L/16 & 224$\times$224 & \capfiltl{}(14M + LAION) & 129.1M & 130.2M & - \\
\midrule
\pevlpre        & \albefl & 256$\times$256 & 14M$\rightarrow$RefCOCO\{,+,g\}+F30KE+GQA+VCR+VG & 14.4M & 15.2M & 4.7M \\
\pevlgrd        & \pevlpre & 512$\times$512 & \pevlpre$\rightarrow$RefCOCO\{,+,g\}+F30KE & 14.4M & 15.2M & 4.7M \\
\pevlvrd        & \pevlpre & 512$\times$512 & \pevlpre$\rightarrow$VG & 14.4M & 15.2M & 6.2M \\
\midrule
\xvlmb          & Swin-B/32 & 224$\times$224 & 4M & 4.0M & 5.1M & 6.2M \\
\xvlml          & Swin-B/32 & 224$\times$224 & 14M + Objects365 + OpenImages & 17.4M & 16.2M & 12.4M \\
\bottomrule
\end{tabular}
    }
    \caption{Overview of core evaluated models. All the models cross-attend to visual features, and use contrastive learning (CL) and a (masked) language modelling objective. Fine-grained models also predict object locations. Unsupervised pretraining data includes COCO~\cite{coco}, SBU~\cite{sbu}, VG~\citep{visual_genome}, CC$_\text{3M}$~\cite{cc3m}, CC$_\text{12M}$~\cite{cc12m} and LAION~\cite{laion}. Supervised data additionally includes RefCOCO and RefCOCO+~\cite{refcoco}, RefCOCOg~\cite{refcocog}, F30KE~\cite{f30k_entities}, GQA~\cite{gqa}, VCR~\cite{vcr}, Objects365~\cite{o365} and OpenImages~\cite{openimages}. In~\cref{tab:models}, we also list downstream performance on VQAv2~\cite{vqav2}, NLVR2~\cite{nlvr2} and RefCOCO+.}
    \label{tab:models_app}
\end{table*}

\subsection{Evaluated Models: Details} \label{app:models}
We provide more details on the models we use to evaluate progress in fine-grained \vl understanding.
See \cref{tab:models_app} for an overview.\footnote{Each model's text and multimodal layers were originally initialised with the weights of BERT$_\text{BASE}$~\cite{bert}.}

\paragraph{}
\noindent\textbf{ALBEF}~\cite{albef} is a recent VLM that has gained popularity due to its design choices, effectively combining core components in \vl learning, such as a contrastive objective and cross-attention, that result in strong downstream performance.
\albef is a dual-stream encoder~\cite{unmasked} that first encodes images and captions independently with a vision~(ViT;~\citealt{vit,deit}) and text~(BERT;~\citealt{bert}) Transformer, respectively; and then fuses them in a cross-modal Transformer.
The model is pretrained with three objectives: masked language modelling (MLM), unimodal image--text contrastive learning and cross-modal image--text matching.
We refer to the original work for more details.
While \albef does not explicitly train for fine-grained understanding, it serves as an important baseline since our three other models build on top of it.

\paragraph{}
\noindent \textbf{BLIP}~\cite{blip} is a unified \vl understanding and generation model, that can be applied to a wide range of downstream tasks.
A key component to \blip's success is \capfilt: a dataset boostrapping method which the authors use to generate synthetic captions and removing noisy pairs from large-scale Web data.
Moreover, unlike any other model we evaluate, \blip uses an autoregressive language modelling (LM) objective to convert visual information into coherent captions, allowing us to evaluate the potential benefits of this objective to learn fine-grained relationships.
\blip is not explicitly trained for fine-grained understanding, however, we believe it is important to assess whether generative language modelling and its data contributions that enhance downstream performance also lead to better fine-grained skills.

\paragraph{}
\noindent\textbf{PEVL}~\cite{pevl} explicitly connects image regions and text tokens through cross-modal position modelling.
Similar to Pix2Seq~\cite{pix2seq_mtl}, \pevl expresses visual positions in text by appending the bounding box coordinates corresponding to a given (annotated) entity in the caption, surrounded by two special tokens `\texttt{<}' and `\texttt{>}': ``\texttt{A cat < 10 73 206 175 > is napping.}''
The bounding box coordinates are discretised and added to the text vocabulary.
Starting from an \albefl checkpoint, \pevl is pretrained by recovering masked text and position tokens through a generalised MLM objective.
The model was trained on a diverse corpus of referring expressions, captions with visual coreferences, question answering, commonsense reasoning, object detection and region descriptions data~(\cref{tab:models}).
Unlike \albef, \pevl is explicitly trained to learn fine-grained, grounded representations of entities by predicting their coordinates in a unified MLM framework.
We evaluate three different models released by the authors, which differ in their pretraining and fine-tuning data: \pevlpre, underwent a second-stage pretraining on multiple supervised tasks (\cref{tab:models_app}); \pevlgrd, which was further fine-tuned for position-output tasks such as phrase grounding~\cite{f30k_entities}; and \pevlvrd, which was fine-tuned for the position-input task of visual relation detection~\cite{visual_genome}.

\paragraph{}
\noindent\textbf{X-VLM}~\cite{x-vlm} also aims at learning to locate visual concepts in the image given the associated texts.
Similar to the \albef architecture, the model consists of an image encoder, a text encoder, and a cross-modal encoder. 
However, unlike \pevl, \xvlm models visual position through an additional bounding box prediction head: given the visually grounded representation of an object label, the model is trained to regress the object's bounding box (bbox) coordinates.
The authors use both object detection labels and region descriptions to learn multi-grained alignments.
The pretraining objective is a linear combination of this bbox loss and the losses defined in \albef to align texts and visual concepts (for more details, see \cref{sec:x-vlm}).

\paragraph{}
In addition to the above models, which we extensively discuss, we also evaluate the following models, based on dual-encoder and frozen LLMs.

\paragraph{}
\noindent\textbf{CLIP}~\cite{clip} is a widely used dual-encoder network.
The model consists of two encoders, one for images and one for text, trained to represent both modalities in a joint space via an unsupervised contrastive objectives over more than 400M image--text pairs from the Web.
Due to its simplicity and wide adoption, we report its performance as a strong, representative baseline.

\paragraph{}
\noindent\textbf{ClipCap}~\cite{clipcap} is an autoregressive encoder--decoder network.
The image encoder is a pretrained CLIP model, while the text decoder is a pretrained GPT-2~\cite{gpt2} language model.
The authors propose to learn a lightweight Transformer-based network to map CLIP embeddings into a fixed length prefix.
The mapping network and the text decoder are fine-tuned to learn how to generate captions, while the CLIP image encoder is frozen.
At inference time, the model generates the caption word after word, starting from the CLIP-based prefix.
We report performance for the two released versions---one fine-tuned on COCO, the other on CC$_\text{3M}$---by ranking positive and negative samples on their likelihood.

\paragraph{}
\noindent\textbf{Flamingo}~\cite{alayrac2022flamingo} is a state-of-the-art VLM capable of tackling a wide range of vision and language tasks from a few input/output examples.
To achieve this, the model relies on a pretrained \clip-like image encoder and a strong pretrained LLM~\citep{chinchilla}, both kept frozen.
To ingest images and videos, the model learns a small fixed number of visual tokens~\citep{set_transformer,perceiver}.
The model is pretrained to generate text from a sequence of text tokens interleaved with images and/or videos.

\paragraph{}
\noindent\textbf{BLIP-2}~\cite{blip2} is the most recent, state-of-the-art VLM based on frozen large image encoders and frozen LLMs~\citep{opt,flan_t5}.
Like \clipcap, \bliptwo learns a mapping network, which in this case is a Transformer model initialised from BERT$_\text{BASE}$.
The mapping network learns visual query tokens to map the visual representations to the frozen LLM in two stages: a \vl representation stage, and a generative learning stage.
The model was pretrained with the same objectives and on the same 129M image--caption data as \blip.
Following the authors' setup for image--text retrieval and matching, we use the \bliptwo model after the first-state pretraining.

\begin{table*}[t!]
    \setlength{\tabcolsep}{2pt}
    \small
    \centering
    \resizebox{\linewidth}{!}{
        \hypersetup{citecolor=lightgray}
\begin{tabular}{rlr|cccccc|cccc}
\toprule
& \multicolumn{2}{c|}{\textbf{Model}} & \multicolumn{1}{c}{\textbf{SVO-Probes}} & \multicolumn{1}{c}{\textbf{VALSE}} & \multicolumn{1}{c}{\textbf{VSR Random}} & \multicolumn{3}{c|}{\textbf{Winoground}} & \multicolumn{2}{c}{\textbf{Flickr30K}} & \multicolumn{2}{c}{\textbf{COCO}} \\
& Name & \multicolumn{1}{c|}{Size} & Avg. & Avg. & Test Avg. & Text & Image & Group &  TR@1    & IR@1   & TR@1   & IR@1  \\
\midrule
\multicolumn{1}{r|}{} & \gray{Random} & \gray{} & \gray{50.0} & \gray{50.0} & \gray{50.0} & \gray{25.0} & \gray{25.0} & \gray{12.5} & \gray{~~0.1} & \gray{~~0.1} & \gray{0.02} & \gray{0.02} \\\midrule
\multicolumn{1}{r|}{} & \gray{LXMERT} & \gray{263M} & \gray{-} & \gray{59.6} & \gray{72.5$^\dagger$} & \gray{19.2} & \gray{~~7.0} & \gray{~~4.0} & \gray{-} & \gray{-} & \gray{-} & \gray{-} \\
\multicolumn{1}{r|}{} & \gray{UNITER$_\text{Large}$} & \gray{303M} & \gray{-} & \gray{-} & \gray{-} & \gray{38.0} & \gray{14.0} & \gray{10.5} & \gray{80.7} & \gray{66.2} & \gray{64.1} & \gray{48.8} \\
\multicolumn{1}{r|}{} & \gray{12-in-1}  & \gray{270M}               & \gray{-} & \gray{75.1} & \gray{-} & \gray{-} & \gray{-} & \gray{-} & \gray{-} & \gray{67.8$^\dagger$} & \gray{-} & \gray{68.0$^\dagger$} \\
\multicolumn{1}{r|}{} & \gray{\clip (ViT-B/32)}  & \gray{151M}                   & \gray{81.6} & \gray{64.0} & \gray{N/A} & \gray{30.7} & \gray{10.5} & \gray{~~8.0} & \gray{88.0} & \gray{68.7} & \gray{58.4} & \gray{37.8} \\
\multicolumn{1}{r|}{} & \gray{\clipcap{}$_\text{CC3M}$} & \gray{295M} & \gray{83.1} & \gray{65.7} & \gray{N/A} & \gray{12.2} & \gray{14.7} & \gray{~~5.5} & \gray{26.4} & \gray{44.1} & \gray{~~6.7} & \gray{24.3} \\
\multicolumn{1}{r|}{} & \gray{\clipcap{}$_\text{COCO}$} & \gray{295M} & \gray{84.1} & \gray{68.5} & \gray{N/A} & \gray{12.2} & \gray{14.7} & \gray{~~5.5} & \gray{27.8} & \gray{52.2} & \gray{~~8.1} & \gray{38.4} \\
\multicolumn{1}{r|}{} & \gray{\flamingo} & \gray{80B} & \gray{88.4} & \gray{\bf 75.3} & \gray{N/A} & \gray{-} & \gray{-} & \gray{-} & \gray{-} & \gray{-} & \gray{-} & \gray{-} \\
\multicolumn{1}{r|}{} & \gray{\bliptwo} & \gray{1.2B} & \gray{86.5} & \gray{74.0} & \gray{61.5} & \gray{43.0} & \gray{22.0} & \gray{18.2} & \gray{\textbf{95.5}} & \gray{\textbf{86.7}} & \gray{\textbf{80.7}} & \gray{\textbf{64.2}} \\

\midrule
\multicolumn{1}{r|}{1} & \albefb  & 500M        & 87.6 & 69.1 & 57.3 & 29.2 & 15.5 & 11.0 & 85.2 & 69.4 & 69.7 & 51.1 \\
\multicolumn{1}{r|}{2} & \xvlmb{}$^\sharp$  & 239M        & 88.9 & 72.4 & 63.0 & 44.0 & \textbf{26.7} & \textbf{21.5} & 85.3 & 71.9 & 70.8 & 55.6 \\
\midrule
\multicolumn{1}{r|}{3} & \albefl  & 500M        & 88.6 & 69.4 & 58.3 & 32.5 & 16.2 & 12.7 & 90.9 & 75.9 & 73.2 & 54.8 \\
\multicolumn{1}{r|}{4} & \blipl  & 638M         & 48.7 & 67.8 & 49.7 & 36.5 & 18.5 & 14.5 & 82.6 & 78.4 & 70.4 & 57.3 \\
\multicolumn{1}{r|}{5} & \pevlpre{}$^\sharp$  & 500M       & 86.2 & 68.9 & 57.5 & 33.2 & 15.7 & 12.2 & 74.9 & 60.0 & 45.9 & 33.2 \\
\multicolumn{1}{r|}{6} & \pevlgrd{}$^\sharp$  & 502M       & 88.5 & 69.5 & 57.7 & 36.2 & 15.0 & 12.0 & 71.8 & 77.6 & 42.8 & 37.7 \\
\multicolumn{1}{r|}{7} & \pevlvrd{}$^\sharp$  & 502M       & 84.8 & 64.5 & 59.5 & 31.2 & 12.0 & ~~7.5 & 68.0 & 55.7 & 38.3 & 30.6 \\
\multicolumn{1}{r|}{8} & \xvlml{}$^\sharp$  & 239M         & \textbf{90.0} & {74.5} & \textbf{64.3} & \textbf{46.7} & 24.5 & 21.2 & 87.7 & 74.9 & 71.6 & 56.1\\
\midrule
\multicolumn{1}{r|}{9} & \blipxl  & 638M         & 51.4 & 68.8 & 46.9 & 35.5 & 15.0 & 11.7 & 90.2 & 79.5 & 71.9 & 58.6 \\
\multicolumn{1}{r|}{10} & \blipxlfilt  & 638M    & 51.2 & 68.2 & 48.7 & 34.7 & 15.2 & 12.2 & 89.1 & 79.7 & 72.2 & 57.8 \\
\multicolumn{1}{r|}{11} & \blipvitxl  & 1.1B     & 50.8 & 70.3 & 50.3 & 34.7 & 14.5 & 12.2 & 90.4 & 80.6 & 74.2 & 59.3 \\
\bottomrule
\end{tabular}
    }
    \caption{Overall performance of our evaluated models on fine-grained benchmarks and zero-shot retrieval tasks. The overall best values for each task are marked in \textbf{bold}. $^\sharp$ marks the fine-grained models. $^\dagger$ denotes performance after task fine-tuning. \xvlm significantly outperforms the other models that we evaluate on fine-grained tasks.}
    \label{tab:overall_baselines_app}
\end{table*}

\begin{table*}[t!]
    \setlength{\tabcolsep}{3.5pt}
    \small
    \centering
    \resizebox{\linewidth}{!}{
        \begin{tabular}{ l c| c| ccc| c| cc| cc| c| c }
        \toprule
        \multirow{2}{*}{\bf Model} &
        \multicolumn{1}{c}{\bf Existence} & \multicolumn{1}{c}{\bf Plurality} & \multicolumn{3}{c}{\bf Counting} & \multicolumn{1}{c}{\bf Sp.rel.$\ddagger$} & \multicolumn{2}{c}{\bf Action} & \multicolumn{2}{c}{\bf Coreference} & \multicolumn{1}{c}{\multirow{2}{*}{\bf Foil-it!}} & \multicolumn{1}{c}{\multirow{2}{*}{\bf Avg.}} \\
        & \multicolumn{1}{c|}{quantifiers} & \multicolumn{1}{c|}{number} & \multicolumn{1}{c}{balanced} & \multicolumn{1}{c}{sns.$\dagger$} & \multicolumn{1}{c|}{adv.$\dagger$} & \multicolumn{1}{c|}{relations} & \multicolumn{1}{c}{repl.$\dagger$} & \multicolumn{1}{c|}{actant swap} & \multicolumn{1}{c}{standard} & \multicolumn{1}{c|}{clean} &
        \multicolumn{1}{c|}{}\\ 
        \midrule
        \multicolumn{1}{l}{\gray{Random}} & \gray{50.0} & \gray{50.0} & \gray{50.0} & \gray{50.0} & \gray{50.0} & \gray{50.0} & \gray{50.0} & \gray{50.0} & \gray{50.0} & \gray{50.0} & \gray{50.0} & \gray{50.0} \\
        \midrule
        \multirow{1}{*}{\gray{GPT-2}} & \gray{58.0} & \gray{51.9} & \gray{51.6} & \gray{49.8} & \gray{45.3} & \gray{75.0} & \gray{66.8} & \gray{76.9} & \gray{54.5} & \gray{50.0} & \gray{80.7} & \gray{60.1} \\
        \midrule
        \multirow{1}{*}{\gray{CLIP}} & \gray{66.9} & \gray{56.2} & \gray{62.1} & \gray{62.5} & \gray{57.5} & \gray{64.3} & \gray{75.6} & \gray{68.6} & \gray{52.1} & \gray{49.7} & \gray{88.8} & \gray{64.0} \\
        \multirow{1}{*}{\gray{LXMERT}} & \gray{78.6} & \gray{64.4} & \gray{62.2} & \gray{69.2} & \gray{42.6} & \gray{60.2} & \gray{54.8} & \gray{45.8} & \gray{46.8} & \gray{44.2} & \gray{87.1} & \gray{59.6} \\
        \multirow{1}{*}{\gray{12-in-1}} & \gray{\bf 95.6} & \gray{72.4} & \gray{\bf 76.7} & \gray{\bf 80.2} & \gray{77.3} & \gray{67.7} & \gray{65.9} & \gray{58.9} & \gray{\bf 75.7} & \gray{\bf 69.2} & \gray{86.9} & \gray{75.1} \\
        \gray{\clipcap{}$_\text{CC3M}$} & \gray{66.3} & \gray{54.8} & \gray{49.4} & \gray{50.1} & \gray{51.5} & \gray{83.2} & \gray{75.5} & \gray{87.9} & \gray{45.1} & \gray{45.2} & \gray{94.7} & \gray{65.7} \\
        \gray{\clipcap{}$_\text{COCO}$} & \gray{74.9} & \gray{60.6} & \gray{55.0} & \gray{53.0} & \gray{53.0} & \gray{\textbf{89.7}} & \gray{71.0} & \gray{86.5} & \gray{47.5} & \gray{49.0} & \gray{\textbf{97.1}} & \gray{68.5} \\
        \gray{\flamingo}   & \gray{63.6} & \gray{59.8} & \gray{58.2} & \gray{55.2} & \gray{\bf 80.2} & \gray{\bf 89.7} & \gray{\bf 86.7} & \gray{\bf 92.8} & \gray{72.2} & \gray{65.4} & \gray{97.0} & \gray{\bf 75.3} \\
        \gray{\bliptwo}    & \gray{83.6} & \gray{\textbf{79.6}} & \gray{70.2} & \gray{68.7} & \gray{68.0} & \gray{65.6} & \gray{84.4} & \gray{63.2} & \gray{62.6} & \gray{58.7} & \gray{96.0} & \gray{74.0} \\
        \midrule
        \albefb         & 71.3 & 78.8 & 62.2 & 65.1 & 59.8 & 73.1 & 73.6 & 58.4 & 52.4 & 55.8 & 95.5 & 69.1 \\
        \xvlmb          & 80.0 & 77.8 & 69.0 & 68.4 & 72.5 & 74.8 & 77.3 & 65.0 & 50.1 & {48.1} & 92.5 & 72.4 \\
        \midrule
        \albefl         & 69.5 & 76.0 & 61.5 & 61.0 & 64.5 & 70.7 & 77.6 & 60.5 & 55.9 & 61.5 & 96.1 & 69.4 \\
        \blipl          & 82.4 & 73.8 & 61.8 & 62.6 & 63.7 & 65.2 & 74.7 & 55.2 & 52.3 & {42.3} & 92.3 & 67.8 \\
        \pevlpre        & 89.7 & 65.5 & 66.0 & 66.2 & 57.3 & 67.9 & 73.5 & 59.4 & 58.2 & 56.7 & 90.9 & 68.9 \\
        \pevlgrd        & 91.1 & 63.9 & 70.0 & 70.9 & 63.2 & 62.4 & 74.4 & 57.1 & 53.8 & {49.0} & 92.6 & 69.5 \\
        \pevlvrd        & 83.8 & 61.8 & 62.8 & 70.3 & {40.4} & 64.5 & 68.1 & 53.2 & {47.7} & {42.3} & 94.1 & 64.5 \\
        \xvlml          & 83.6 & 78.7 & 71.5 & 72.0 & 74.8 & 73.1 & 79.2 & 64.6 & 60.0 & {49.0} & 91.9 & 74.5 \\
        \midrule
        \blipxl         & 78.2 & 75.9 & 63.4 & 63.4 & 58.5 & 66.2 & 75.2 & 59.0 & 56.4 & 52.9 & 93.2 & 68.8 \\
        \blipxlfilt     & 75.4 & 75.0 & 64.7 & 68.8 & 53.0 & 66.7 & 73.0 & 60.6 & {48.2} & 51.0 & 93.8 & 68.2 \\
        \blipvitxl      & 73.3 & 77.7 & 68.2 & 67.6 & 61.2 & 71.8 & 75.3 & 60.8 & 51.1 & {45.2} & 96.1 & 70.3 \\
        \bottomrule
    \end{tabular}

    }
    \vspace{-2mm}
    \caption{Performance on the VALSE benchmark according to pairwise ranking accuracy. Best results are in \textbf{bold}.\\
    $\dagger${\bf sns.} Counting small numbers. {\bf adv.} Counting adversarial. {\bf repl.} Action replacement. $\ddagger$ {\bf Sp.rel.} Spatial relations.}
    \label{tab:valse}
\end{table*}

\begin{table*}[t!]
    \setlength{\tabcolsep}{3pt}
    \small
    \centering
    \resizebox{\textwidth}{!}{
        \begin{tabular}{lrrr|rrr|rrr|rrr|rrr}
\toprule
{\bf Model} & \multicolumn{3}{c}{{\bf Object}} & \multicolumn{3}{c}{{\bf Relation}} & \multicolumn{3}{c}{{\bf Both}} & \multicolumn{3}{c}{{\bf 1 Main Pred}} & \multicolumn{3}{c}{{\bf 2 Main Preds}} \\
& Text & Image & Group & Text & Image & Group & Text & Image & Group & Text & Image & Group & Text & Image & Group \\
\midrule
 \gray{Random}                  & \gray{25.00} & \gray{25.00} & \gray{12.50} & \gray{25.00} & \gray{25.00} & \gray{12.50} & \gray{25.00} & \gray{25.00} & \gray{12.50} & \gray{25.00} & \gray{25.00} & \gray{12.50} & \gray{25.00} & \gray{25.00} & \gray{12.50} \\
 \gray{MTurk Human}                  & \gray{92.20} & \gray{90.78} & \gray{88.65} & \gray{89.27} & \gray{90.56} & \gray{86.70} & \gray{76.92} & \gray{57.69} & \gray{57.69} & \gray{87.33} & \gray{85.62} & \gray{82.53} & \gray{95.37} & \gray{96.30} & \gray{93.52} \\
\midrule
 \gray{LXMERT}                       & \gray{22.70}          & \gray{9.22}           & \gray{6.38}           & \gray{17.60}          & \gray{5.58}           & \gray{2.58}           & \gray{15.38}          & \gray{7.69}           & \gray{3.85}           & \gray{19.18}          & \gray{8.56}           & \gray{5.14}           & \gray{19.44}          & \gray{2.78}           & \gray{0.93}           \\
 \gray{UNITER$_\text{Large}$}             & \gray{39.01} & \gray{12.77}          & \gray{9.93}           & \gray{36.05} & \gray{14.16}          & \gray{9.87}           & \gray{50.00} & \gray{19.23}          & \gray{19.23} & \gray{40.07} & \gray{16.44}          & \gray{13.36}          & \gray{32.41} & \gray{7.41}           & \gray{2.78}           \\
 \gray{\clip (ViT-B/32)}              & \gray{34.75} & \gray{7.80}           & \gray{6.38}           & \gray{22.75}          & \gray{8.58}           & \gray{5.58}           & \gray{\bf 80.77} & \gray{\bf 42.31} & \gray{\bf 38.46} & \gray{35.27} & \gray{13.01}          & \gray{10.27}          & \gray{18.52}          & \gray{3.70}           & \gray{1.85}           \\
 \gray{\clipcap{}$_\text{CC3M}$}               & \gray{14.18}          & \gray{17.02}          & \gray{7.80}           & \gray{11.16}          & \gray{12.02}          & \gray{3.43}           & \gray{11.54}          & \gray{26.92} & \gray{11.54}          & \gray{13.70}          & \gray{16.10}          & \gray{6.51}           & \gray{8.33}           & \gray{11.11}          & \gray{2.78}           \\
 \gray{\clipcap{}$_\text{COCO}$}             & \gray{12.77}          & \gray{17.02}          & \gray{5.67}           & \gray{12.88}          & \gray{9.87}           & \gray{3.86}           & \gray{23.08}          & \gray{34.62} & \gray{19.23} & \gray{14.73}          & \gray{16.44}          & \gray{6.85}           & \gray{10.19}          & \gray{7.41}           & \gray{1.85}           \\
 \gray{\bliptwo}               & \gray{47.52} & \gray{\bf 27.66} & \gray{\bf 21.99} & \gray{38.20} & \gray{17.60}          & \gray{14.59}          & \gray{61.54} & \gray{30.77} & \gray{30.77} & \gray{48.63} & \gray{26.37} & \gray{22.26} & \gray{27.78} & \gray{10.19}          & \gray{7.41}           \\
 \midrule
 \albefb                & {29.79} & 12.77          & 8.51           & {26.61} & 15.02          & 10.73          & {50.00} & {34.62} & {26.92} & {33.22} & 19.18          & 14.04          & 18.52          & 5.56           & 2.78           \\
 \xvlmb                 & {46.10} & {\bf 27.66} & {\bf 21.99} & {41.63} & {\bf 24.46}          & {19.31} & {53.85} & {\bf 42.31} & {\bf 38.46} & {47.60} & {\bf 30.48} & {25.68} & {34.26} & {\bf 16.67}          & {\bf 10.19}          \\
 \midrule
 \albefl                & {29.79} & 15.60          & 9.22           & {30.90} & 14.16          & 12.02          & {61.54} & {38.46} & {\bf 38.46} & {35.27} & 18.49          & 14.38          & 25.00          & 10.19          & 8.33           \\
 \blipl              & {41.13} & 24.11          & {17.73} & {32.19} & 14.16          & 11.16          & {50.00} & {26.92} & {26.92} & {42.12} & 21.92          & {18.15} & 21.30          & 9.26           & 4.63           \\
 \pevlpre  & {31.21} & 14.89          & 10.64          & {33.48} & 14.59          & 11.59          & {42.31} & {30.77} & {26.92} & {36.30} & 19.52          & 15.75          & 25.00          & 5.56           & 2.78           \\
 \pevlgrd & {39.01} & 14.89          & 12.77          & {33.91} & 13.73          & 10.30          & {42.31} & {26.92} & {23.08} & {37.67} & 17.47          & 15.07          & {32.41} & 8.33           & 3.70           \\
 \pevlvrd       & {26.95} & 10.64          & 7.09           & {32.19} & 12.45          & 6.87           & {46.15} & 15.38          & 15.38          & {31.85} & 11.64          & 8.22           & {29.63} & 12.96          & 5.56           \\

 \xvlml                & {\bf 48.23} & 23.40          & {19.86} & {\bf 44.21} & 23.18          & {\bf 20.17} & {61.54} & {\bf 42.31} & {\bf 38.46} & {\bf 51.03} & {29.11} & {\bf 26.03} & {\bf 35.19} & 12.04          & 8.33           \\
 \midrule
 \blipxl             & {37.59} & 17.02          & 10.64          & {34.76} & 12.02          & 10.73          & {30.77} & {30.77} & {26.92} & {40.07} & 18.84          & 14.73          & 23.15          & 4.63           & 3.70           \\
 \blipxlfilt & {34.04} & 16.31          & 11.35          & {33.48} & 13.30          & 11.16          & {50.00} & {26.92} & {26.92} & {38.70} & 19.18          & 15.41          & 24.07          & 4.63           & 3.70           \\
 \blipvitxl       & {35.46} & 16.31          & 13.48          & {32.62} & 12.88          & 11.59          & {50.00} & 19.23          & 11.54          & {39.04} & 17.81          & 15.07          & 23.15          & 5.56           & 4.63           \\
\bottomrule
\end{tabular}

    }
    \vspace{-2mm}
    \caption{Results on Winoground by linguistic tag. Best results are in \textbf{bold}.}
    \label{tab:winoground_ling}
    \vspace{4mm}

    \begin{minipage}{.6\linewidth}
        \scriptsize
        \centering
        \resizebox{\textwidth}{!}{
            \begin{tabular}{lrrr|rrr|rrr}
\toprule
{\bf Model} & \multicolumn{3}{c}{{\bf Symbolic}} & \multicolumn{3}{c}{{\bf Pragmatics}} & \multicolumn{3}{c}{{\bf Same Image Series}} \\
& Text & Image & Group & Text & Image & Group & Text & Image & Group \\
\midrule
 \gray{Random}                  & \gray{25.00} & \gray{25.00} & \gray{12.50} & \gray{25.00} & \gray{25.00} & \gray{12.50} & \gray{25.00} & \gray{25.00} & \gray{12.50} \\
 \gray{MTurk Human}                  & \gray{96.43} & \gray{92.86} & \gray{92.86} & \gray{58.82} & \gray{41.18} & \gray{41.18} & \gray{95.65} & \gray{91.30} & \gray{91.30} \\
 \midrule
 \gray{LXMERT}                       & \gray{28.57} & \gray{3.57}           & \gray{3.57}           & \gray{17.65}          & \gray{5.88}           & \gray{0.00}           & \gray{8.70}           & \gray{4.35}           & \gray{0.00}           \\
 \gray{UNITER$_\text{Large}$}             & \gray{39.29} & \gray{28.57} & \gray{17.86} & \gray{35.29} & \gray{0.00}           & \gray{0.00}           & \gray{4.35}           & \gray{8.70}           & \gray{0.00}           \\
 \gray{\clip (ViT-B/32)}              & \gray{39.29} & \gray{3.57}           & \gray{3.57}           & \gray{35.29} & \gray{5.88}           & \gray{5.88}           & \gray{8.70}           & \gray{0.00}           & \gray{0.00}           \\
 \gray{\clipcap{}$_\text{CC3M}$}               & \gray{21.43}          & \gray{21.43}          & \gray{10.71}          & \gray{5.88}           & \gray{5.88}           & \gray{0.00}           & \gray{0.00}           & \gray{8.70}           & \gray{0.00}           \\
 \gray{\clipcap{}$_\text{COCO}$}             & \gray{25.00}          & \gray{25.00}          & \gray{14.29}          & \gray{23.53}          & \gray{17.65}          & \gray{\bf 17.65} & \gray{13.04}          & \gray{13.04}          & \gray{0.00}           \\
 \gray{\bliptwo}               & \gray{42.86} & \gray{28.57} & \gray{25.00} & \gray{41.18} & \gray{\bf 23.53}          & \gray{\bf 17.65} & \gray{21.74}          & \gray{13.04}          & \gray{4.35}           \\
 \midrule
 \albefb                 & {42.86} & 25.00          & {17.86} & 17.65          & 17.65          & 5.88           & 8.70           & 0.00           & 0.00           \\
 \xvlmb                 & {50.00} & {\bf 32.14} & {\bf 32.14} & {41.18} & {\bf 23.53}          & {\bf 17.65} & {30.43} & {\bf 26.09} & {\bf 13.04}          \\
 \midrule
 \albefl                & {39.29} & 14.29          & 14.29          & 17.65          & 0.00           & 0.00           & {26.09} & 4.35           & 4.35           \\
 \blipl              & {39.29} & 25.00          & {17.86} & 23.53          & 17.65          & {\bf 17.65} & 8.70           & 4.35           & 0.00           \\
 \pevlpre  & {35.71} & 14.29          & 14.29          & {29.41} & 11.76          & 5.88           & 13.04          & 8.70           & 4.35           \\
 \pevlgrd & {35.71} & 7.14           & 7.14           & {29.41} & 11.76          & 11.76          & {26.09} & 8.70           & 4.35           \\
 \pevlvrd       & {42.86} & 10.71          & 7.14           & 23.53          & 5.88           & 0.00           & {\bf 34.78} & 17.39          & 8.70           \\
 \xvlml                & {42.86} & 21.43          & {17.86} & {\bf 47.06} & 11.76          & 5.88           & {26.09} & 4.35           & 4.35           \\
 \midrule
 \blipxl             & {\bf 57.14} & 14.29          & 14.29          & {35.29} & 11.76          & 11.76          & {26.09} & 0.00           & 0.00           \\
 \blipxlfilt & {50.00} & 14.29          & 14.29          & {35.29} & 5.88           & 5.88           & 21.74          & 0.00           & 0.00           \\
 \blipvitxl       & {39.29} & 14.29          & 14.29          & {29.41} & 0.00           & 0.00           & 13.04          & 0.00           & 0.00           \\
 \bottomrule
\end{tabular}

        }
        \vspace{-2mm}
        \caption{Results on Winoground by visual tag. Best results are in \textbf{bold}.}
        \label{tab:winoground_vis}

    \end{minipage}%
    \hfill
    \begin{minipage}{.38\linewidth}
        \scriptsize
        \centering
        \resizebox{\textwidth}{!}{
            \begin{tabular}{lrrrr}
\toprule
\textbf{Model}                & \textbf{ Subj.}  & \textbf{Verb} & \textbf{Obj.} & \textbf{Avg.} \\
\midrule
\gray{Random} & \gray{50.0} & \gray{50.0} & \gray{50.0} & \gray{50.0} \\
\midrule
\gray{\clip (ViT-B/32)} & \gray{83.6} & \gray{79.0} & \gray{88.1} & \gray{81.6} \\
\gray{\clipcap{}$_\text{CC3M}$} & \gray{84.2} & \gray{80.5} & \gray{90.2} & \gray{83.1} \\
\gray{\clipcap{}$_\text{COCO}$} & \gray{87.3} & \gray{81.5} & \gray{89.8} & \gray{84.1} \\
\gray{\flamingo} & \gray{90.1} & \gray{86.7} & \gray{92.3} & \gray{88.4} \\
\gray{\bliptwo} & \gray{87.6} & \gray{84.6} & \gray{91.7} & \gray{86.5} \\
\midrule
\albefb & 88.5 & 85.4 & 93.7 & 87.6 \\
\xvlmb & 89.3 & 87.1 & 94.5 & 88.9 \\
\midrule
\albefl & 89.4 & 86.4 & 94.7 & 88.6 \\
\blipl & 49.8 & 48.8 & 47.5 & 48.7 \\
\pevlpre & 89.4 & 82.9 & 93.9 & 86.2 \\
\pevlgrd & \textbf{91.2} & 85.9 & 94.6 & 88.5 \\
\pevlvrd & 90.1 & 81.1 & 92.3 & 84.8 \\
\xvlml & 90.3 & \textbf{88.4} & \textbf{94.6} & \textbf{90.0} \\
\midrule
\blipxl & 50.8 & 51.4 & 51.8 & 51.4 \\
\blipxlfilt & 49.4 & 51.3 & 52.5 & 51.2 \\
\blipvitxl & 50.0 & 50.9 & 50.9 & 50.8 \\
\bottomrule
\end{tabular}
        }
        \vspace{-2mm}
        \caption{Performance on the SVO-Probes benchmark according to pairwise ranking accuracy. Best results are in \textbf{bold}.}
        \label{tab:svo}
        \vspace{4mm}
    \end{minipage}
    
    \small
    \centering
    \resizebox{\linewidth}{!}{
        \begin{tabular}{lcccccccc}
\toprule
\textbf{Model} & \textbf{Adjacency} & \textbf{Directional} & \textbf{Orientation} & \textbf{Projective} & \textbf{Proximity} & \textbf{Topological} & \textbf{Unallocated} & \textbf{Overall} \\
\midrule
\gray{Random} & \gray{50.0 / 50.0} & \gray{50.0 / 50.0} & \gray{50.0 / 50.0} & \gray{50.0 / 50.0} & \gray{50.0 / 50.0} & \gray{50.0 / 50.0} & \gray{50.0 / 50.0} & \gray{50.0 / 50.0} \\
\midrule
\gray{\bliptwo}        & \gray{59.8 / 54.9} & \gray{50.0 / 43.3} & \gray{52.5 / 57.1} & \gray{59.8 / 63.6} & \gray{56.2 / 51.2} & \gray{66.4 / 67.0} & \gray{75.0 / 66.7} & \gray{61.2 / 61.5} \\
\midrule
\albefb         & 52.3 / 51.1 & 38.6 / 42.2 & 55.9 / \textbf{58.0} & 61.7 / 60.2 & 56.2 / \textbf{55.3} & 58.6 / 59.2 & 65.6 / 56.9 & 58.0 / 57.3 \\
\xvlmb          & 57.6 / 57.7 & 56.8 / 43.3 & 59.3 / 52.7 & \textbf{69.2} / 66.1 & 57.8 / 54.5 & \textbf{71.2} / 68.4 & 75.0 / 62.7 & 66.6 / 63.0 \\
\midrule
\albefl         & 52.3 / 54.2 & 59.1 / 40.0 & 55.9 / \textbf{58.0} & 59.8 / 62.6 & 46.9 / 52.0 & 66.8 / 58.9 & 71.9 / 58.8 & 60.2 / 58.3 \\
\blipl          & 56.8 / 49.3 & 56.8 / 50.0 & 57.6 / 47.3 & 42.5 / 49.3 & 51.6 / 48.0 & 45.1 / 51.8 & 50.0 / 41.2 & 47.4 / 49.7 \\
\pevlpre        & 47.0 / 55.3 & 56.8 / 48.9 & 57.6 / 56.2 & 61.9 / 60.8 & 51.6 / 48.8 & 62.4 / 57.4 & 71.9 / 58.8 & 59.3 / 57.5 \\
\pevlgrd        & 53.8 / 53.5 & \textbf{65.9} / 50.0 & 59.3 / 52.7 & 60.9 / 59.4 & 60.9 / 54.5 & 62.7 / 60.2 & 75.0 / 58.8 & 61.1 / 57.7 \\
\pevlvrd        & 54.5 / 55.6 & 59.1 / 52.2 & 61.0 / 53.6 & 59.8 / 60.4 & 59.4 / 54.5 & 64.1 / 63.1 & 68.8 / 64.7 & 60.7 / 59.5 \\
\xvlml          & \textbf{61.4} / \textbf{58.5} & \textbf{65.9} / 46.7 & \textbf{64.4} / \textbf{58.0} & 68.4 / \textbf{67.7} & \textbf{62.5} / 52.0 & 70.5 / \textbf{68.7} & \textbf{84.4} / \textbf{68.6} & \textbf{67.9} / \textbf{64.3} \\
\midrule
\blipxl         & 44.7 / 41.2 & 43.2 / 52.2 & 52.5 / 53.6 & 53.6 / 45.4 & 53.1 / 49.6 & 50.2 / 49.7 & 40.6 / 37.3 & 50.5 / 46.9 \\
\blipxlfilt     & 57.6 / 49.3 & 36.4 / 57.8 & 47.5 / 53.6 & 45.9 / 45.5 & 48.4 / 47.2 & 48.5 / 51.1 & 37.5 / 41.2 & 47.7 / 48.7 \\
\blipvitxl      & 56.1 / 51.8 & 29.5 / \textbf{58.9} & 49.2 / 52.7 & 46.9 / 48.5 & 53.1 / 43.9 & 49.8 / 51.8 & 46.9 / 47.1 & 48.7 / 50.3 \\
\bottomrule
\end{tabular}

    }
    \vspace{-2mm}
    \caption{Dev/Test results on the VSR Random dataset. Best results are in \textbf{bold}.}
    \label{tab:vsr}
\end{table*}

\subsection{Re-implementation Setup} \label{app:reimplement}
We re-implement ALBEF and X-VLM in JAX~\citep{dm_jax} to ensure full control of modelling, data, and initialisation decisions.\footnote{To verify our implementation, we compare an \albef model trained in our codebase with one trained in the original codebase. Specifically, we pretrain both models on COCO by initialising their visual encoder with a CLIP ViT-B/16 model, and their text encoder with a BERT$_\text{BASE}$ model. The two models perform similarly on both zero-shot Flickr30K and COCO retrieval tasks with a gap below 1pp Recall@1.}
We note \albef's vision encoder is initialised with a pretrained ViT-B/16 encoder~\citep{deit} with an input resolution of 256$\times$256 pixels, but \xvlm adopts a more efficient Swin-B/32~\cite{swin} encoder with input resolution of 224$\times$224 pixels.
In our re-implementation we initialise both models with a ViT-B/16 with a 224$\times$224 input resolution pretrained on ImageNet-21k~\cite{vit_augreg}, to ensure that different initialisation is not responsible for the results.

\begin{figure*}[t!]
    \centering
    \includegraphics[width=\textwidth, trim={0cm 0cm 0cm 0cm}, clip]{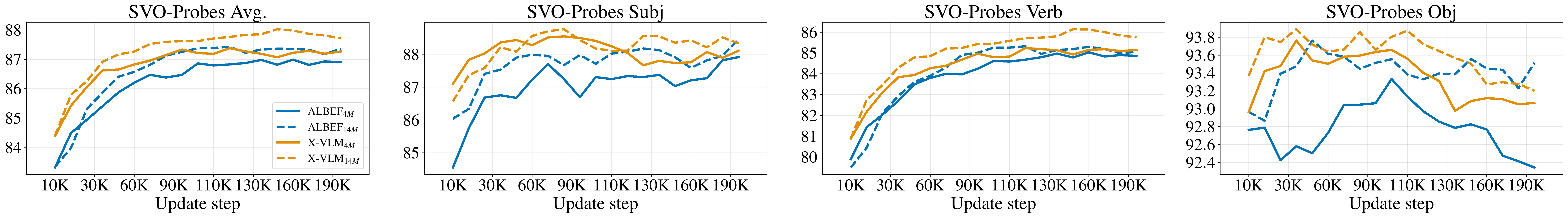}
    \vspace{-7mm}
    \caption{Training dynamics on SVO-Probes subtasks. Random performance is 50\%.}\label{fig:svo_dynamics}
\end{figure*}

\begin{figure*}[t!]
    \centering
    \includegraphics[width=\textwidth, trim={0cm 0cm 0cm 0cm}, clip]{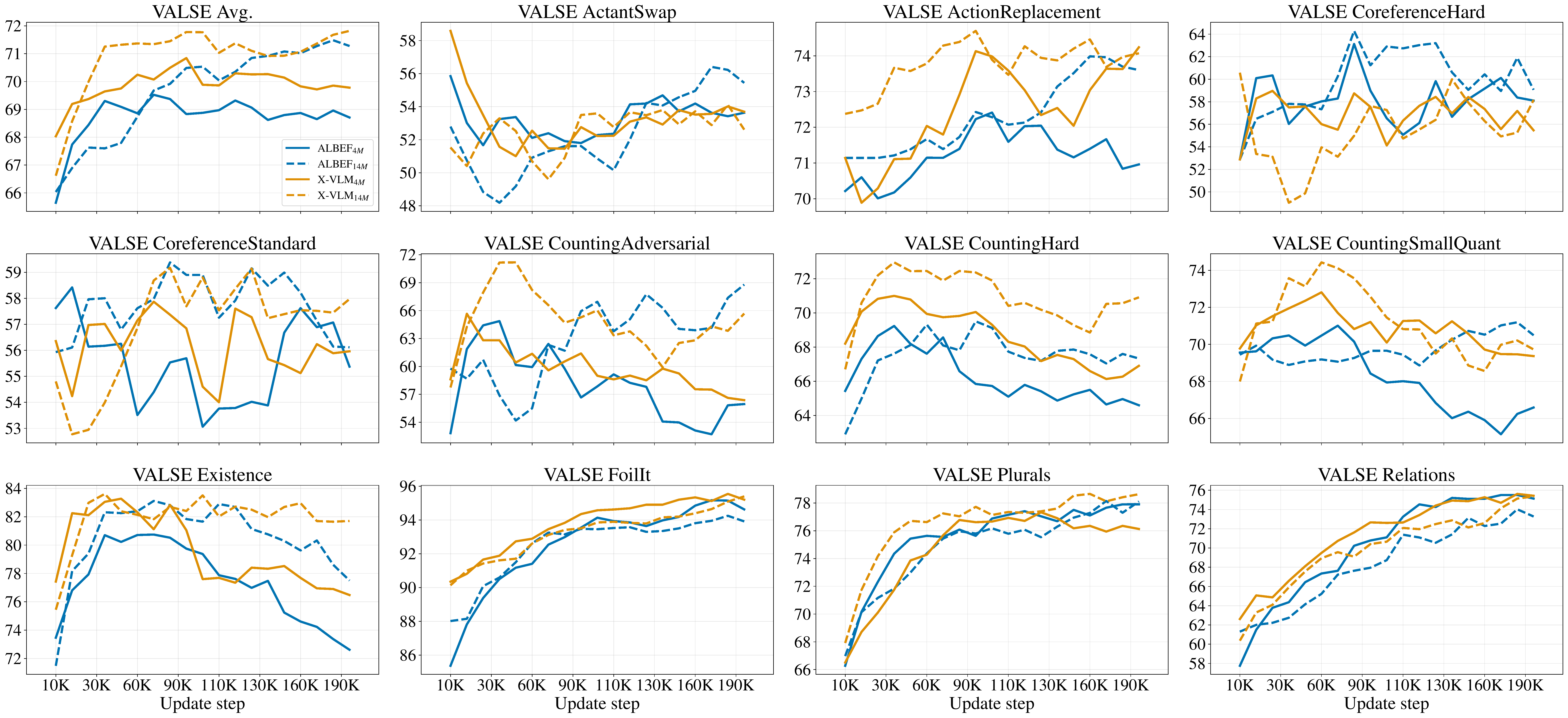}
    \vspace{-7mm}
    \caption{Training dynamics on VALSE subtasks. Random performance is 50\%.}\label{fig:valse_dynamics}
\end{figure*}

We pretrain our models on the same 4M and 14M datasets that were originally used by the authors~(\cref{tab:models}), but note that only 1.8M and 11.2M data points were available for CC$_\text{3M}$ and CC$_\text{12M}$, respectively.
For object detection data, we use the data points released by the X-VLM authors, and interleave captioning and detection data with a 2:1 ratio following their official implementation.
Following \cite{x-vlm}, we pretrain our models for 200K steps using a batch size of 512 and 1024 samples for \albef and \xvlm, respectively. We pretrain once, using the same hyperparameters as the authors.\footnote{The \xvlm authors trained for 200K steps of image captioning data, not counting batches of detection datasets. We count each batch towards the final number of steps, hence effectively training for fewer steps than~\citet{x-vlm}.}
Training our models takes around 1.5 days on Cloud TPUv4 (a 2x2x2 slice).
We evaluate our models on both fine-grained benchmarks (SVO-Probes, VALSE and VSR) and on two zero-shot, coarse retrieval tasks (Flickr30K and COCO).

\section{Results}\label{app:results}

\subsection{Results by Subtask} \label{app:res_subtasks}
\cref{tab:overall_baselines_app} compares overall performance of our evaluated models~(\cref{sec:eval_models}) with the state-of-the-art models in each of four fine-grained benchmarks~(\cref{sec:benchmarks}).
Results for each subtask are reported in Tables~\cref{tab:winoground_vis,tab:winoground_ling,tab:svo,tab:valse,tab:vsr}.

In addition to the core discussion in~\cref{sec4}, we note that \flamingo achieves the overall best performance on VALSE; and that the coarse-grained \bliptwo model performs remarkably well on our range of fine-grained tasks, especially on VALSE, VSR and Winoground.
This could be due to a number of factors, such as a larger ViT encoder, the usage of visual queries and the different formulations for the ITC and ITM objectives.
We leave a deeper investigation of large VLMs to future work.

Moreover, we also note that \clipcap well on VALSE \texttt{spatial relations} and \texttt{action} subtasks, wherein its GPT-2 backbone already performs better than most VLMs.
This is further proof of the efficacy of adapting strong LMs for \vl tasks.

\subsection{Full Dynamics of Fine-grained Tasks} \label{app:res_dynamics}

\begin{figure*}[t!]
    \centering
    \includegraphics[width=0.88\textwidth, trim={0cm 0cm 0cm 0cm}, clip]{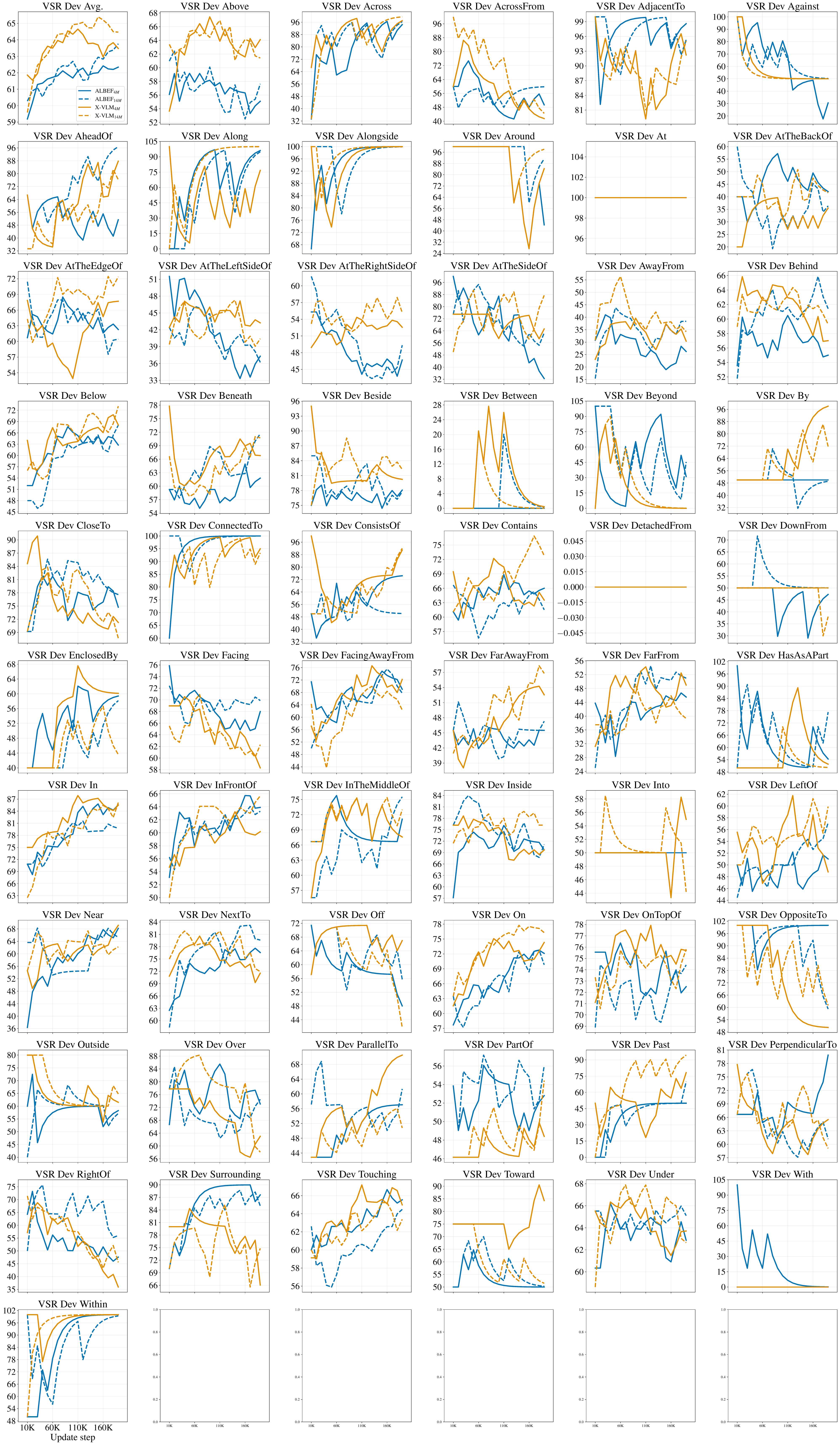}
    \vspace{-3mm}
    \caption{Training dynamics on VSR Random dev set subtasks. Random performance is 50\%.}\label{fig:vsr_dev_dynamics}
\end{figure*}

\begin{figure*}[t!]
    \centering
    \includegraphics[width=0.88\textwidth, trim={0cm 0cm 0cm 0cm}, clip]{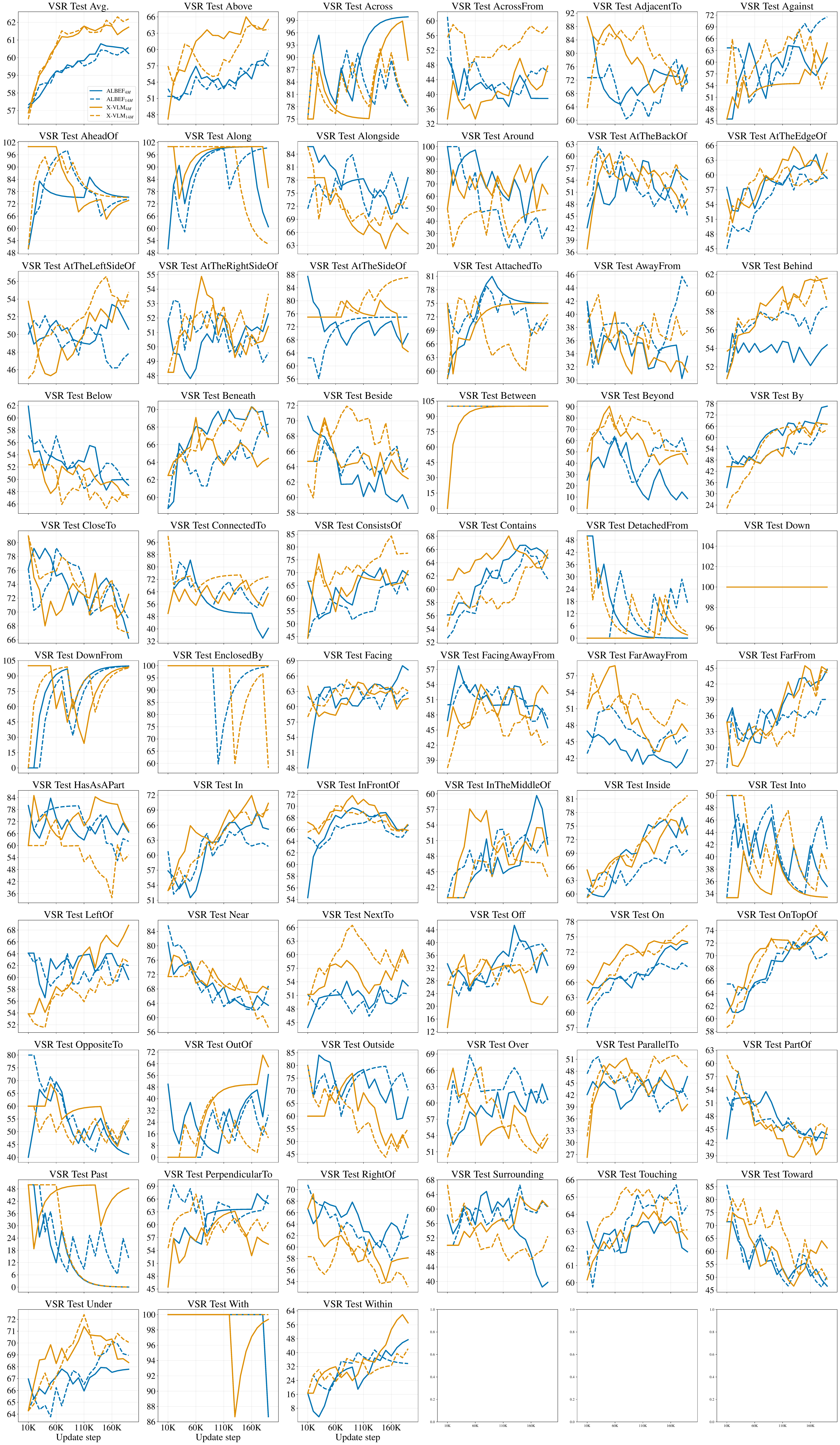}
    \vspace{-3mm}
    \caption{Training dynamics on VSR Random test set subtasks. Random performance is 50\%.}\label{fig:vsr_test_dynamics}
\end{figure*}

Figures\cref{fig:svo_dynamics,fig:valse_dynamics,fig:vsr_dev_dynamics,fig:vsr_test_dynamics} display pretraining dynamics for our re-implemented \albefb, \albefl, \xvlmb, and \xvlmm models. 
For better visualisation, our curves have been smoothed by a 0.6 factor through exponential moving average.

\begin{figure*}[t!]
    \centering
    \includegraphics[width=\textwidth, trim={0cm 0cm 0cm 0cm}, clip]{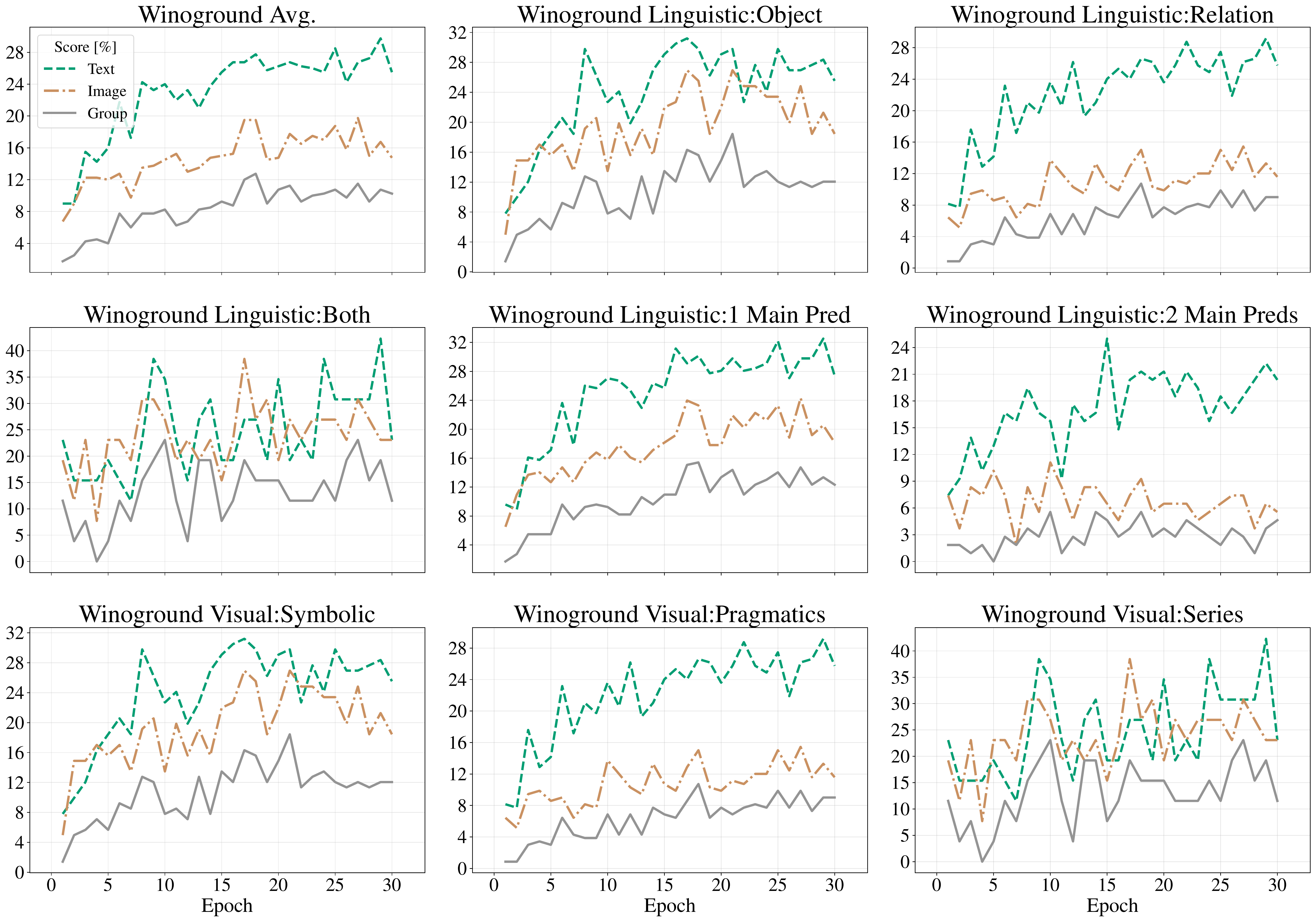}
    \vspace{-7mm}
    \caption{Training dynamics on Winoground subtasks of \albefb pretrained with the official codebase on GCP. Random performance is 25\% for Text Score and Image Score, and 16\% for Group Score.}\label{fig:albef_winoground}
    \vspace{20cm}
\end{figure*}

Finally, \cref{fig:albef_winoground} shows how performance on Winoground evolves when pretraining an \albefb model.\footnote{We note that we used an image resolution of 224$\times$224 pixels, and a batch size of 256 (instead of 512) as we pretrained on a GCP instance with 4$\times$ A100 GPUs (instead of the 8$\times$ A100 GPUs originally used by the authors).}
Looking at overall performance, we see that a model's score can vary by more than 4pp from one epoch to the next.
While longer pretraining seems beneficial, some subtasks, such as \texttt{Linguistic:Both} and \texttt{Visual:Series}, fluctuate considerably; and after 20 epochs, the Image score starts decreasing on other subtasks, such as \texttt{Linguistic:Object} and \texttt{Visual:Symbolic}.

\end{document}